\documentclass{article}

\usepackage[final]{neurips_2024}

\usepackage[utf8]{inputenc} % allow utf-8 input
\usepackage[T1]{fontenc}    % use 8-bit T1 fonts
\usepackage{hyperref}       % hyperlinks
\usepackage{url}            % simple URL typesetting
\usepackage{booktabs}       % professional-quality tables
\usepackage{amsfonts}       % blackboard math symbols
\usepackage{amsmath}
\usepackage{multirow}
\usepackage{pifont}
\usepackage{multicol}
\usepackage{wrapfig}
\usepackage[capitalize,noabbrev]{cleveref}
\usepackage{nicefrac}       % compact symbols for 1/2, etc.
\usepackage{microtype}      % microtypography
\usepackage{xcolor}         % colors
\usepackage{graphicx}
\usepackage{algorithm}
\usepackage{algpseudocode}
\usepackage{subcaption}
\usepackage{amsthm}
\usepackage{enumitem}
\usepackage{bbm}

\usepackage{listings} % for lstlisting environment
\newcommand*{\affaddr}[1]{#1} % No op here. Customize it for different styles.
\newcommand*{\affmark}[1][*]{\textsuperscript{#1}}

\newtheorem{theorem}{Theorem}

\newtheorem{lemma}{Lemma}
\newtheorem{corollary}{Corollary}[theorem]

\theoremstyle{definition}
\newtheorem{definition}{Definition}

\theoremstyle{remark}

\newcommand{\myparagraph}[1]{\vspace{0pt}\noindent{\bf #1}}

\newcommand{\ie}{\textit{i.e.}}

\newcommand{\cmark}{\ding{51}}%
\newcommand{\xmark}{\ding{55}}%
\DeclareMathOperator{\softmax}{softmax}
\DeclareMathOperator{\depth}{depth}

\newcommand{\erank}[1]{\operatorname{\rho}(#1)}

\newcommand{\norm}[1]{\left\lVert#1\right\rVert}

\title{DeNetDM: Debiasing by Network Depth Modulation}

\author{%
{Silpa Vadakkeeveetil Sreelatha\thanks{Equal contribution.} \hspace{0.1pt}
\affmark[1]}
\hspace{1pt}
Adarsh Kappiyath\footnotemark[1]  
\hspace{0.1pt}   
\affmark[1]
\hspace{1pt}
Abhra Chaudhuri\affmark[1,2,3]
\hspace{1pt}
Anjan Dutta\affmark[1]\\
\affaddr{\affmark[1] University of Surrey}
\hspace{1pt}
\affaddr{\affmark[2] University of Exeter}
\hspace{1pt}
\affaddr{\affmark[3] Fujitsu Research of Europe}\\
\texttt{\scriptsize \{s.vadakkeeveetilsreelatha, a.kappiyath, anjan.dutta\}@surrey.ac.uk, abhra.chaudhuri@fujitsu.com} 
}

\begin{document}

\maketitle

% Abstract
\begin{abstract}
Neural networks trained on biased datasets tend to inadvertently learn spurious correlations, hindering generalization. We formally prove that (1) samples that exhibit spurious correlations lie on a lower rank manifold relative to the ones that do not; and (2) the depth of a network acts as an implicit regularizer on the rank of the attribute subspace that is encoded in its representations. Leveraging these insights, we present DeNetDM, a novel debiasing method that uses network depth modulation as a way of developing robustness to spurious correlations. Using a training paradigm derived from Product of Experts, we create both biased and debiased branches with deep and shallow architectures and then distill knowledge to produce the target debiased model. Our method requires no bias annotations or explicit data augmentation while performing on par with approaches that require either or both. We demonstrate that DeNetDM outperforms existing debiasing techniques on both synthetic and real-world datasets by 5\%. The project page is available at \href{https://vssilpa.github.io/denetdm/}{https://vssilpa.github.io/denetdm/}.
\end{abstract}

% Introduction
\section{Introduction}
\label{sec:intro}
Deep neural networks (DNNs) have made remarkable progress across various domains by delivering superior performance on large-scale datasets. However, while the benefits of training DNNs on large-scale datasets are undeniable, these algorithms also tend to inadvertently acquire unwanted biases \cite{NEURIPS2020_6cfe0e61}, hampering their generalization. For instance, a classifier predominantly trained to recognize camels in desert landscapes could encounter difficulties when attempting to identify a camel situated on a road \cite{Kim_2021_ICCV}. While a certain degree of bias can enhance model performance, as exemplified by the assumption that cars usually travel on roads \cite{Choi_2020_CVPR}, it remains critical to identify and address unwanted biases. 
% \begin{figure}
%     \centering
%     \includegraphics[width=\textwidth]{figures/teaser_diagram.pdf}
%     \caption{Our DeNetDM model comprises a Product of Expert (PoE) \cite{Hinton:02}, featuring one deep and one shallow expert. The shallow expert primarily captures core attributes, achieving debiased classification, while the deep expert focuses on bias. To compensate for the shallow expert's limited capability due to less depth, we implement a knowledge transfer strategy utilizing knowledge from the experts to train a target debiased model.}
%     \label{fig:teaser-diagram}
% \end{figure}
Previous methods to address this problem rely on bias annotations as suggested in \cite{9607491,Kim_2019_CVPR,sagawa2019distributionally,wang2020fair}, and may involve predefined bias types, such as texture bias mitigation approach in \cite{geirhos2018imagenettrained}. 
% . For instance, in \cite{geirhos2018imagenettrained}, texture bias is mitigated by training a shape-oriented classifier using augmented data and style transfer. However, this approach limits the ability of models to achieve robustness against biases for which obtaining prior knowledge is a challenging endeavor. 
However, acquiring bias labels with human resources is expensive and time-consuming. Recent studies, including \cite{NEURIPS2020_eddc3427} and \cite{NEURIPS2021_disentangled},  have shifted towards debiasing methods without bias labels, with approaches like \cite{NEURIPS2020_eddc3427} emphasizing bias-aligned samples and reweighting bias-conflicting samples, while others like \cite{NEURIPS2021_disentangled,Kim_2021_ICCV} introduce augmentation strategies to diversify bias-conflicting points.

We propose DeNetDM (\textbf{De}biasing by \textbf{Net}work \textbf{D}epth \textbf{M}odulation), a novel approach to automatically identify and mitigate spurious correlations in image classifiers without relying on explicit data augmentation or reweighting. We start by showing that a sample set that exhibits bias through spurious correlation of attributes lies on a manifold with an effective dimensionality (rank) lower than its bias-free counterpart.
% We then leverage this finding to formally derive a relationship between the depth of a network and the true rank of the attribute (not sample) subspace that it learns and generalizes to.
We then leverage this finding to formally derive a relationship between the depth of a network and the true rank of the attribute (not sample) subspace that it encodes. We find for a set of attributes that are equally likely to minimize the empirical risk, a deeper network prefers to retain those with a lower rank, with a higher probability. This implies that the depth of a network acts as an implicit regularizer in the rank space of the attributes.
We find that deeper networks tend to generalize based on bias attributes and shallower networks tend to generalize based on core attributes. This finding is in line with a number of works that show that deeper networks tend to learn low rank solutions in general \citep{roy2007effrank, huh2023simplicitybias, wang2024implicit}.
% Note, however, that previous works make no claim about the relationship between the depth of a network and the attribute subspace that it generalizes to, a link we establish in our work for the first time, to the best of our knowledge.
Note, however that prior works do not establish the relationship between network depth and the rank of the attribute subspace, a link we establish in our work for the first time, to the best of our knowledge.

Our theoretical claims are confirmed by our preliminary empirical study on linear feature decodability, which quantifies the extent to which specific data attributes can be accurately and reliably extracted from a given dataset or signal. Our study focuses on the feature decodability of bias and core attributes in the neural networks of varying depths, following the approach outlined in \cite{NEURIPS2020_71e9c662}. Our observations in untrained neural networks reveal that the feature decodability tends to diminish as the networks become deeper. We also investigate how attribute decodability varies with Empirical Risk Minimization (ERM) based training on networks of varying depths. 
Our hypothesis posits that in a task requiring deep and shallow branches to acquire distinct information, the deep branch consistently prioritizes bias attributes, while the shallow branch favors core attributes. We utilize a technique inspired by the Product of Experts \citep{Hinton:02}, where one expert is deeper than the other. Empirical analysis shows that the deep branch becomes perfectly biased and the shallow branch becomes relatively debiased by focusing solely on the core attributes by the end of the training. Since the shallow branch may lack the capacity to capture the nuances of the core attributes adequately due to less depth, we propose a strategy where we train a deep debiased model utilizing the information acquired from both deep (perfectly biased) and shallow (weak debiased) network in the previous phase. Our training paradigm efficiently facilitates the learning of core attributes from bias-conflicting data points to the debiased model of any desired architecture. 

In summary, we make the following contributions: (1) We theoretically prove that the deep models prefer to learn spurious correlations compared to shallower ones, supported by empirical analysis of the decodability of bias and core attributes across neural networks of varying depths. (2) Building upon the insights from our decodability experiments, we present a novel debiasing approach that involves training both deep and shallow networks to obtain a desired debiased model. (3) We perform extensive experiments and ablation studies on a diverse set of datasets, including synthetic datasets like Colored MNIST and Corrupted CIFAR-10, as well as real-world datasets, Biased FFHQ, BAR and CelebA, demonstrating an approximate 5\% improvement over existing methods.
    % \item Our results demonstrate that our approach outperforms existing state-of-the-art methods, with an approximate 5\% performance improvement.

% Related Works
\section{Related Works}
\label{sec:related_works}
Several works, such as \cite{NEURIPS2020_71e9c662,10.1145/3457607}, have highlighted neural networks' vulnerability to spurious correlations during empirical risk minimization training. Recently, various debiasing techniques have emerged, which can be categorized as follows.

\myparagraph{Supervision on bias:} A variety of approaches (e.g., \cite{9607491,Kim_2019_CVPR,sagawa2019distributionally,wang2020fair}) assume readily accessible bias labels for bias mitigation. Some approaches assume prior knowledge on specific bias types without using explicit annotations, like texture bias in \cite{wang2018learning,NEURIPS2021_e5afb0f2,geirhos2018imagenettrained}. Recent works such as \cite{karimi-mahabadi-etal-2020-end, clark-etal-2019-dont} apply the Product of Experts method to mitigate bias in natural language processing, assuming a biased expert's availability. However, obtaining bias labels can be resource-intensive. In contrast, DeNetDM, our proposed method, does not require pre-access to bias labels or types. Instead, it leverages diverse network architecture depths within the Product of Experts framework to implicitly capture relevant bias and core attributes.

\myparagraph{Utilization of pseudo bias-labels:} Recent approaches avoid explicit bias annotations by obtaining pseudo-labels through heuristics to identify biased samples. One heuristic suggests that biases easy to learn are captured early in training, as seen in \cite{NEURIPS2020_eddc3427, NEURIPS2021_disentangled, liu2023avoiding, Kim_2021_ICCV, pmlr-v202-tiwari23a, 10.1609/aaai.v37i12.26748}. \cite{NEURIPS2020_eddc3427} employ generalized cross-entropy loss to identify and reweight bias-conflicting points. On the other hand, \cite{NEURIPS2021_disentangled} augment features of bias-conflicting points for debiasing, while \cite{liu2023avoiding} employ logit correction and group mixup techniques to diversify bias-conflicting samples. Other methods like \cite{NEURIPS2020_e0688d13} and \cite{Seo_2022_CVPR} acquire pseudo-bias labels through clustering in biased network feature spaces. Our approach does not explicitly require pseudo-bias labels; it implicitly uses them during training to learn both biased and debiased models. 

\myparagraph{Dependence on network architectures:} \cite{diffenderfer2021a} employ lottery-ticket-style pruning algorithms for compressed robust architectures. Similarly, approaches like \cite{DBLP:conf/cvpr/ParkLLY23,pmlr-v139-zhang21a} introduce pruning to extract robust subnetworks.  Our method aligns with this category but does not target specific robust subnetwork discovery. Instead, we utilize training dynamics of varied-depth architectures to enhance debiasing. Meanwhile, \cite{shrestha2022occamnets} applies Occam's razor principle to optimize network depth and visual regions, enhancing overall robustness. Both DeNetDM and OccamNets \citep{shrestha2022occamnets} aim to simplify learning for better generalization and reduced spurious correlations. DeNetDM uses depth modulation with separate deep and shallow branches to address bias -- where the shallow model captures biases and the deep model learns complex, unbiased patterns. In OccamNets, simplicity is a core design principle, with the architecture adaptively minimizing complexity on a per-sample basis. Both methods tackle spurious correlations without extra annotations or data augmentation but through distinct architectural strategies.

% % Method
\begin{figure*}[!t]
  \centering
  \includegraphics[width=\textwidth]{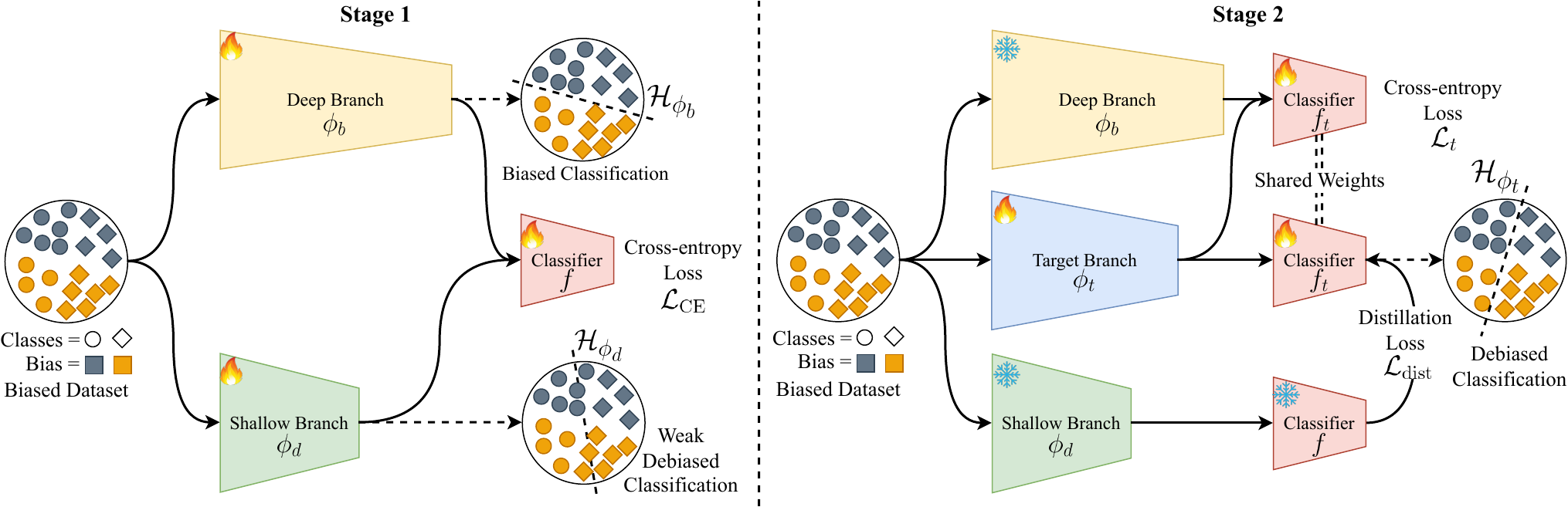}  % Use one of the provided example images
  \caption{\textbf{Illustration of the DeNetDM framework}: In Stage 1, an ensemble of shallow and deep branches produces outputs linearly combined and trained as a product of experts. The cross-entropy loss with depth modulation aids in separating biases and identifying target attributes. In Stage 2, we further introduce a target branch with the desired architecture, which also requires debiasing. This phase exclusively focuses on refining the target branch's feature extractor ($\phi_{t}$) and classifier head ($f_{t}$) while leveraging knowledge from the initial stages.}
  \label{fig:block_diagram}
\end{figure*}

\section{Debiasing by Network Depth Modulation}
\label{sec:DenetDM}
First, we theoretically justify that the deeper models are more inclined to learn spurious correlations compared to shallow networks, as discussed in \cref{sec:theory}. We then provide empirical evidence to support our theoretical claims by utilizing feature decodability, detailed in \cref{sec:linear_decodability_analysis}. Based on these, we introduce DeNetDM, a debiasing approach centered on network depth modulation.
% We propose DeNetDM, a debiasing approach based on network depth modulation, inspired by the findings discussed in \cref{sec:linear_decodability_analysis}. 
Our training process comprises two stages: initially, a deep and shallow network pair is trained using a training paradigm that originates from
% we train a pair of deep and shallow networks using a 
Products of Experts \citep{Hinton:02}, yielding both biased and debiased models, which is detailed in \cref{sec:poe_training}.
Subsequently, recognizing the limitations of the shallow debiased model in capturing core feature complexities due to its depth, we proceed to train a target debiased model, ensuring it possesses the same or higher depth compared to the deep biased model. This phase leverages information acquired from the biased and debiased models in the previous step, as elaborated in \cref{sec:target_debiased_training}. An illustration of DeNetDM is provided in \cref{fig:block_diagram}.

% \abhra{
\myparagraph{Notations:}
% We operate on a dataset $X$, where $\eta$ fraction of the data points, denoted with $X_a$, are bias-aligned and the remaining $(1 - \eta)$ points, denoted with $X_c$, are bias conflicting.
We operate on a dataset $X$, where a fraction of the data points, denoted with $X_a$, are bias-aligned and the remaining points, denoted with $X_c$, are bias conflicting.
% Let $\phi_n: X \to \mathbb{R}^n$ \anjan{should be $\mathbb{R}$}
% \anjan{I would change one of the $n$s, also in the later sections we are using $\phi_b$ and $\phi_d$ to respectively denote biased and debiased branches, I think an alternative notation would be useful}
% be an encoder of depth $n$ \change{(\ie, $\texttt{depth}(\phi)=n$)} that produces an embedding $z \in \mathbb{R}^n$ for an input $x \in X$.
Let $\phi: X \to \mathbb{R}^n$ be an encoder that produces an embedding $z \in \mathbb{R}^n$ for an input $x \in X$.
We denote the effective rank \citep{roy2007effrank} of a matrix $A$ as $\rho(A)$, which gives us a continuous notion of the size of the span (rank) of $A$, a quantity that is maximized under equally distributed singular values, and minimized when a single singular value dominates over the rest \citep{huh2023simplicitybias}. Let $B$ and $C$ be the set of bias and core attributes respectively, both with strictly positive ranks, defining bases that are orthogonal to each other, \ie, $B \perp C$. A summary of notations is provided in \cref{sec:notations}.
% }

% \abhra{
\subsection{Simplicity Bias and Spurious Correlations}
% }

\label{sec:theory}

% \abhra{
% We start by showing that bias-conflicting data points are inherently difficult to learn than those that are bias-aligned. This leads us to a formal understanding of the relationship between the depth $n$ of an encoder $\phi_n$ \anjan{please make the depth notation consistent with sec. 3.3} and the nature of the subset of $X$ (bias-aligned or bias-conflicting) that it learns with lower generalization error.
% This leads us to understanding of the relationship between the depth of a neural encoder and the nature of the subset of $X$ (bias-aligned or bias-conflicting) that it learns with lower generalization error.
Debiasing with network depth modulation requires understanding how the depth of a neural network affects its learning of bias-aligned or bias-conflicting subsets of $X$ with lower generalization error.
These results finally let us build up to our finding that deeper networks are more susceptible to learning spurious features over their shallower counterparts. All proofs are deferred to \cref{sec:proofs}.
% }

% \abhra{
\begin{definition}[Stability]
    A partitioning $X = X_1 \cup X_2... \cup X_m$ of a sample set $X$ is stable \textit{wrt.} an attribute $\omega$ when:
    \begin{equation*}
        P(X_i^\omega) = P(X^\omega); \forall i \in [1, m],
    \end{equation*}
    % where $S^\omega$ is the subspace of the sample set $S$ corresponding to the attribute $\omega$, and $P(\cdot)$ is the associated probability distribution.
    where $X^\omega$ and $X_i^\omega$ are the respective subspaces of $X$ and $X_i$ corresponding to the attribute $\omega$, and $P(\cdot)$ is the associated probability distribution.
\end{definition}
% }

% \abhra{
% In other words, the distribution of $\omega$ in each of the partitions $X_i$ is an invariant under the partitioning operation, and hence, is the same as the distribution of the full sample set $X$.
For example, if $\omega$ follows a uniform distribution in $X$, a stable partitioning would ensure that each of the partitions $X_i$ also have $\omega$ distributed uniformly.
Stability ensures that a partitioning does not introduce sampling bias into any of the partitions \textit{wrt.} a particular attribute.
% }

% \anjan{is it possible to provide a toy example to guide the readers better understand the above definition?}

% \abhra{
\begin{theorem}[Partition Rank]
    % % Let $\beta_a$ and $\beta_c$ be the respective variances in the core attributes in $X_a$ and $X_c$ respectively.
    % Let $C$ be the random variable corresponding to the core attributes in $X$.
    % When the partitioning $X = X_a \cup X_c$ is stable \textit{wrt.} $C$, the effective rank of the bias-aligned partition is strictly lower than the effective rank of the bias-conflicting partition, \ie,
    % \begin{equation*}
    %     \erank{X_a} < \erank{X_c}
    % \end{equation*}
    When the partitioning $X = X_a \cup X_c$ is stable \textit{wrt.} $C$, the rank of the bias-aligned partition is upper-bounded by the rank of the bias-conflicting partition, \ie,
    \begin{equation*}
        \operatorname{rank}(X_a) \leq \operatorname{rank}(X_c)
    \end{equation*}    
    
    \label{thm:pr}
\end{theorem}
% }
% \abhra{
\textit{Intuition}:
    The theorem assumes a stable partitioning of the sample set X. It implies that, in both the bias-aligned and conflicting subsets, the distribution of the core attributes are equal to that of the original sample set, \ie, $P(X_a^C) = P(X_c^C) = P(X^C)$. Under this condition, the only component in either of the subsets that determines the subset's rank should be the bias attributes, assuming (without loss of generality) that the attribute space is made up of only the core and the bias attributes. The proof proceeds by establishing that the rank of the bias attributes is lower in the bias-aligned points (resulting from the lack of intra-class variation due to spurious correlation with the class label) than in the bias-conflicting points. 
% }

% \abhra{
% \textit{Proof Sketch}:
%     The theorem assumes a stable partitioning of the sample set X, \ie, in both the bias-aligned and conflicting subsets, the distribution of the core attributes are equal to that of the original sample set, \ie,
%     \begin{equation*}
%         P(X_a^C) = P(X_c^C) = P(X^C)
%     \end{equation*}
%     Under this condition, the only component in either of the subsets that determines the subset's rank should be the bias attributes, assuming (without loss of generality) that the attribute space is made up of only the core and the bias attributes. Now, within $X_a$, the variance of the bias attributes is very low by definition, since they all consistently exhibit the same bias. This in turn leads to a low effective rank of the subspace $X_a^B$.  On the other hand, for $X_c$, again by definition, the bias attributes do not occur consistently due to the bias-conflicting nature of the subset, which leads to a higher overall variance, and consequently, a higher effective rank of the subspace $X_c^B$. As argued before, under the core attribute stability assumption, since the effective rank is solely determined by the bias attribute subspace, the effective rank of the bias-aligned partition is lower than that of the one that is bias-conflicting.
% }

% \abhra{
\begin{theorem}[Depth-Rank Duality]
    \label{thm:drd}
    Let $\mathcal{A} = [ A_0, A_1, ..., A_n ]$ be the attribute subspace of $X$ with increasing ranks, \ie, $\operatorname{rank}(A_0) < \operatorname{rank}(A_1) < ... < \operatorname{rank}(A_n)$,
    such that every $A \in \mathcal{A}$ is maximally and equally informative of the label $Y$, \ie, $I(A_0, Y) = I(A_1, Y) = ... = I(A_n, Y)$.
    Then, across the depth of the encoder $\phi$, SGD yields a parameterization that optimizes the following objective:
    \begin{equation}
    \underbrace{\min_{\phi, f} \operatorname{\mathcal{L}}(f(\phi(X)), Y)}_\text{ERM} + \min_{\phi} \sum_d \norm{ \phi[d](\Tilde{X}) - \Omega^d \odot \mathcal{A}}_2,
    \label{eqn:drd_opt}
    \end{equation}
    where 
    $\mathcal{L}(\cdot, \cdot)$ is the empirical risk, $f(\cdot)$ is a classifier head, $\phi[i](\cdot)$ is the output of the encoder $\phi$ (optimized end-to-end) at depth $d$, $\norm{\cdot}_2$ is the $l^2$-norm, $\odot$ is the element-wise product,
    $\Tilde{X}$ is the $l_2$-normalized version of $X$,
    $\Omega^d = [\mathbbm{1}_{\pi_1(d)}; \mathbbm{1}_{\pi_2(d)}; ...; \mathbbm{1}_{\pi_n(d)}]$,
    $\mathbbm{1}_\pi$ is a random binary function that outputs $1$ with a probability $\pi$, and $\pi_i(d)$ is the propagation probability of $A_i$ at depth $d$ bounded as:
    % \begin{equation}
    %     \pi_i(d) = \mathcal{O}(\erank{\phi[d]}^{-1} r_i^{-d}),
    %     \label{eqn:pos}
    % \end{equation}
    \begin{equation}
        \pi_i(d) = \mathcal{O}\left( \erank{\phi[d]} \, r_i^{-d} \right),
        \label{eqn:pos}
    \end{equation}
    where $\erank{\phi[d]}$ is the effective rank of the $\phi[d]$ representation space, and $r_i = \operatorname{rank}(A_i)$.
\end{theorem}
% }

% \abhra{
\textit{Intuition}:
    For a set of attributes, all of which equally minimize the training loss, \cref{thm:drd} describes the strategy adopted by SGD to parameterize a neural encoder, for capturing the above set of attributes. At a given depth $d$ of the encoder $\phi$ (represented as $\phi[d]$), each attribute $A_i \in \mathcal{A}$ gets encoded in the representation space of $\phi[d]$ according to its corresponding probability mass $\pi_i(d)$.
    According to \cref{eqn:pos}, the probability of survival of all attributes decrease with increasing depth. However, the probability of survival of an attribute with a higher rank drops faster with increasing depth than that of one with a lower rank, prioritizing the usage of lower rank attributes at greater depths. In other words, the depth of a network acts as an implicit regularizer in the attribute rank space. 
    % }
    
    % \abhra{
    As an example, say, a neural network $\phi$ of depth $d$ (denoted as $\phi[d]$) has $3K$ available dimensions, and of depth $D > d$, $\phi[D]$ has $K$ available dimensions (the rank reduction with increasing depth stemming from the simplicity bias \citep{huh2023simplicitybias, wang2024implicit}). Say the attribute space it has to learn from is composed of two attributes: (1) $A_0$, with a rank of $K$, and (2) $A_1$, with a rank of $K+i$, where $1 \leq i \leq K$, where both $A_0$ and $A_1$ are equal minimizers of the empirical risk. So, according to \cref{thm:drd}, at $\phi[d]$, the encoder has no constraint over the number of attributes it can accommodate, since $3K \geq 2K + i$. However, at depth $D$,
    % \anjan{are you meaning depth or output at depth $D$? please adjust the notation, also check the next $\phi_D$ at the end of this paragraph, I think it should be $\phi[D]$},
    $\phi[D]$ can only choose an attribute with $K$ dimensions. Since both $A_0$ and $A_1$ result in the same solution for ERM (Empirical Risk Minimization), SGD would parameterize $\phi[D]$ to capture $A_0$ with a higher probability. 

\begin{figure}[!t]
    \centering
    \begin{subfigure}{0.49\textwidth}
        \includegraphics[width=\textwidth]{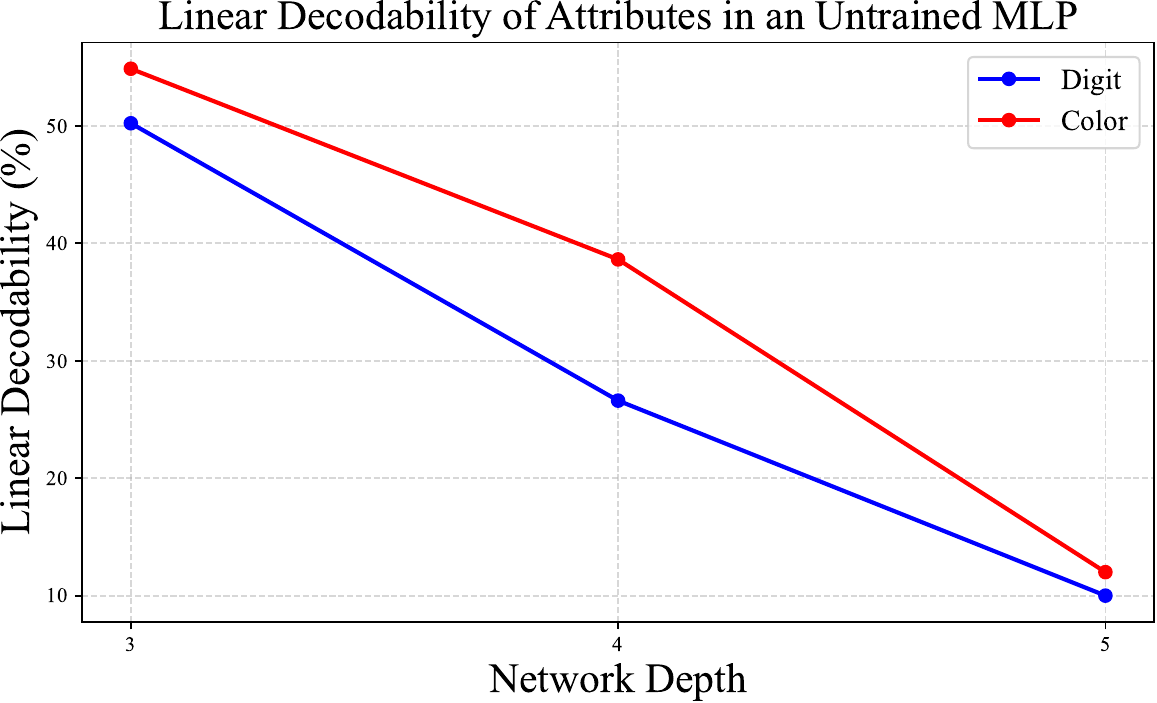}
        \caption{Linear Decodability vs. Network Depth}
        % Variation of linear decodability of features with an increase in network depth shows a clear decreasing trend.}
        \label{fig:linear_decodability_untrained}
    \end{subfigure}
    \hfill
    \begin{subfigure}{0.49\textwidth}
        \includegraphics[width=\textwidth]{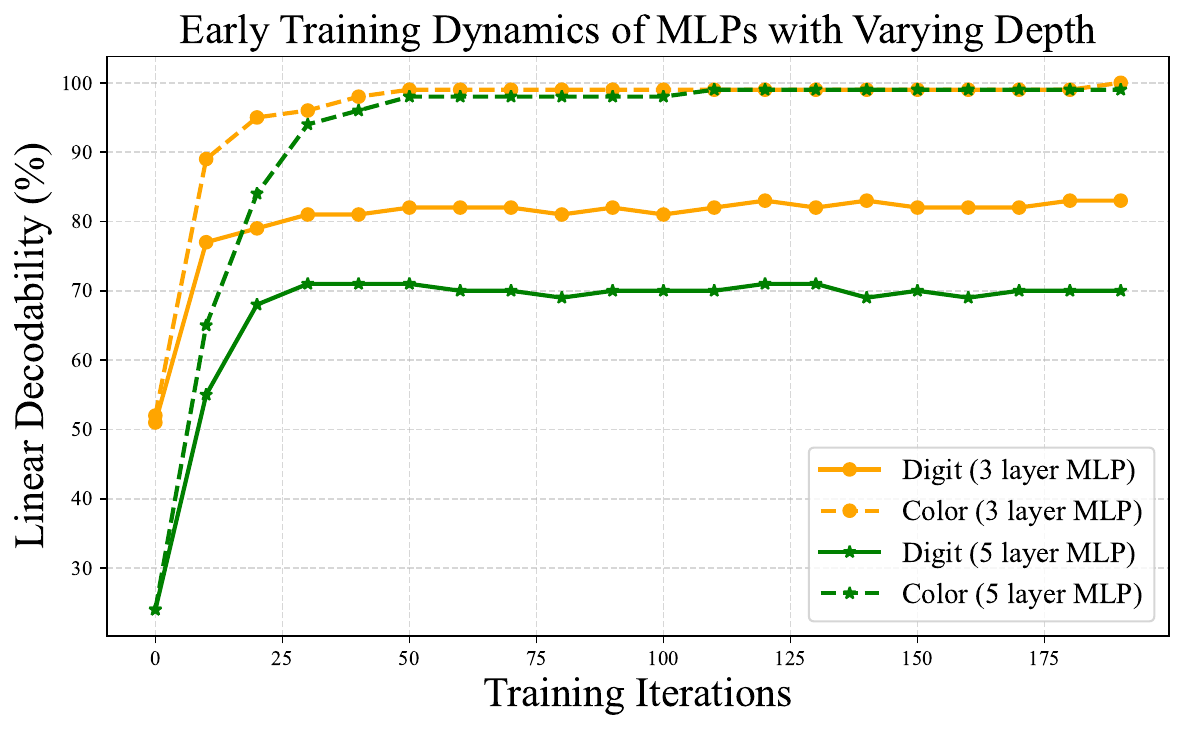}
        \caption{Linear Decodability vs. Training Iterations}
        % Training dynamics measured in terms of linear decodability of features on MLPs of different depths trained using ERM.}
        \label{fig:training_dynamics_erm}
    \end{subfigure}
    \caption{\textbf{Exploring the effect of depth modulation:} (a) illustrates how the linear decodability of features decreases as neural network depth increases, while (b) dives into the training dynamics of MLPs with varying depths under ERM.}
    \label{fig:depth_effect}
\end{figure}

\subsection{Effect of Depth Modulation}
\label{sec:linear_decodability_analysis}
\cref{thm:drd} establishes a relationship between the depth of a network and the nature of the features it learns in terms of its rank. To empirically validate this, we probe MLPs of depths 3, 4, and 5, using the feature decodability technique proposed by \cite{NEURIPS2020_71e9c662}, to uncover the types of features that get encoded in them.  We use the Colored MNIST dataset \cite{NEURIPS2020_eddc3427} (CMNIST), where digit identity (core attribute) is spuriously correlated with color (bias attribute). We experiment with the decodability of the digit identity and color attributes in the CMNIST dataset. Additional information on the computation of feature decodability can be found in \cref{sec:feature_decodability}. We regard digit identity to have a higher rank than that of color, due to its higher representational complexity / information content in terms of the number of bits required for storage, a notion also confirmed in the experiements of \cite{NEURIPS2020_71e9c662}. We start by looking at the decodabilities at random initialization of the networks, and interestingly observe in \cref{{fig:linear_decodability_untrained}} that the decodabilities of both attributes decrease with increasing depth, but that of digit identity drops faster than color. Since at random initialization, there is no notion of empirical risk, the $\min \mathcal{L}(\cdot, \cdot)$ term in \cref{thm:drd} is cancelled out. Thus, the observation aligns with our prediction of the second term in $\norm{\cdot}_2$ of \cref{thm:drd} that the higher the rank of a feature, the less likely it is to get encoded in the later layers, the theoretically predicted behavior specifically for random networks being discussed in \cref{cor:random_drd}. We then proceed to investigating how feature decodability evolves during the early stages of Empirical Risk Minimization (ERM) training across the networks of varying depths, \ie, under the presence of $\min \mathcal{L}(\cdot, \cdot)$, the results of which are summarized in \cref{fig:training_dynamics_erm}. We perform similar linear decodability analysis on C-CIFAR10 dataset and the observations are presented in \cref{supp_sec:decodability_cifar}.

As observed in \cref{fig:training_dynamics_erm}, the initial phases of training for both networks emphasize color attribute (since bias is easy to learn), leading to notable improvements in color decodability for both models. Also, as training progresses, the 3-layer model exhibits higher digit decodability compared to the 5-layer model. Hence, the difference in decodability between color and digit attributes becomes more pronounced in the 5-layer compared to the 3-layer MLP.
This again confirms the prediction of our \cref{thm:drd} that when two attributes equally minimize the empirical risk, a deeper network is more likely to select the one with a lower rank, while a shallower network will try to accommodate as much of both as possible. Based on these observations, the deep models may prefer bias attributes, while shallow models focus on core attributes when tasked with capturing distinct information.

This prompts us to explore whether similar behavior can be induced in models of equal depth. In this scenario, both models, undergoing ERM training, may exhibit a similar trend, with the disparity in decodability between biased and core attributes becoming nearly identical in both models due to same depth.  Consequently, when compelling each model to learn distinct information, they may capture biased or core attributes, or even divide attribute information between them, leading to a loss of control over the bias identification process. We also present empirical evidence in \cref{tab:effect-depth-difference} to support these claims. Therefore, using models of different depths introduces an inductive bias suitable for the bias identification process.

\subsection{Stage1: Segregation of Bias \& Core Attributes}
\label{sec:poe_training}

% In this section, we explain the training procedure to obtain the \textbf{b}iased and \textbf{d}ebiased classifier for an $M$ class classification problem, \abhra{based on our theoretical findings in \cref{thm:pr,thm:drd}}.
\cref{thm:pr} predicts that bias-aligned points lie on a lower-rank manifold than bias conflicting points. \cref{thm:drd} predicts that as we go deeper into a neural network, the likelihood that a higher rank feature, that equally minimizes the empirical risk as that of other lower rank features, is retained, decays exponentially with depth. Based on this, we present a training procedure to obtain the \textbf{b}iased and \textbf{d}ebiased classifier for an $M$ class classification problem.
Let $\phi_b$ and $\phi_d$ denote the parameters of the feature extractors associated with the deep and shallow branches, where $\depth(\phi_b) > \depth(\phi_d)$. We use $f$ to represent the classifier head shared by $\phi_b$ and $\phi_d$. Here, $f$, $\phi_b$ and $\phi_d$ are trainable parameters. Considering an image-label pair $(x, y)$, the objective function is expressed as:
\begin{equation}
  \mathcal{L}_{\text{CE}}(\hat{p}, y) = -\sum_{c=1}^M y_c \log(\hat{p}_c) \label{eq:cross_entropy_stage1}
\end{equation}
where $\hat{p} = \softmax\left(f(\alpha_b \phi_b(x) + \alpha_d \phi_d(x))\right)$. If we set $\alpha_b = \alpha_d = 1$ throughout the training process, we get:
\begin{equation}
  \hat{p} = \softmax\left(f\left(\phi_b(x) + \phi_d(x)\right)\right)
  \label{eq: p_stage1}
\end{equation}
%
% where $\alpha_b$ and $\alpha_d$ are set to 1 throughout the training process. 
To evaluate the performance of an individual expert, we assign a value of 1 to the corresponding $\alpha$ while setting the other $\alpha$ equal to 0.

Our training methodology is derived from the Products of Experts technique \citep{Hinton:02} where multiple experts are combined to make a final prediction, and each expert contributes to the prediction with a weight. However, in our approach, the role of the experts is assumed by $\phi_b$ and $\phi_d$, whose features are combined through weighted contributions. The conjunction of features is then passed to the shared classifier to generate predictions. We provide a detailed proof elucidating the derivation of \cref{eq: p_stage1} through the Product of Experts in \cref{sec:poe_proof} of the Appendix.
Due to the architectural constraints we imposed by modulating their capacities, the deep expert tends to prioritize the learning of bias attribute, while the shallow expert is inclined towards the core attribute. The model leverages the strengths of both experts to effectively learn from their combined knowledge. We investigate the training dynamics in \cref{subsec: training dynamics analysis}. 

\subsection{Stage2 : Training the Target Debiased Model}
\label{sec:target_debiased_training}

The initial phase effectively separates the bias and core attributes into deep and shallow branches, respectively. However, relying solely on the debiased shallow branch may not be practical, as it might not capture the complex features representing the core attributes, given the less depth of the shallow model. This limitation does not apply to the deep biased model. To tackle this challenge, we introduce a target branch with the desired architecture for debiasing.

Let $\phi_{t}$ be the parameters of the feature extractor associated with the target branch and $f_{t}$ be the classifier head whose weights are initialized using the weights of $f$. During this phase, our training is exclusively focused on $\phi_{t}$ and $f_{t}$. We freeze $\phi_b$ and $\phi_d$ since we leverage these models to only extract the necessary knowledge for debiasing the target branch. To capture information orthogonal to $\phi_b$, we employ the same training approach described in \cref{sec:poe_training}, where $\phi_b$ and $\phi_{t}$ serve as the experts. The objective function can be written as:
\begin{equation*}
    \mathcal{L}_t(\hat{p}, y) = -\sum_{c=1}^My_c\log(\hat{p}_c)
\end{equation*}
where
\begin{minipage}{0.94\textwidth}
    \begin{equation}
        \hat{p} = \softmax(f_t(\beta_b \phi_b(x) \\
                + \beta_t \phi_t(x)))
    \label{eq:poe_target}
    \end{equation}    
\end{minipage}

The training and evaluation of the experts follow the procedure described in \cref{sec:poe_training}, with the key difference being that in this phase, only a single expert, $\phi_t$, which is the target branch and classifier $f_t$, undergoes updates. 

We further leverage the knowledge pertaining to the core attributes, which is encapsulated in $\phi_d$, by transferring this knowledge to the target branch $\phi_t$ through knowledge distillation. Here, $\phi_t$ acts as the student, whereas $\phi_d$ corresponds to the teacher. We set $\beta_b=0$ and $\beta_t=1$ in \cref{eq:poe_target} to obtain the predictions of the student $\phi_t$. Therefore, the distillation loss is given by :
\begin{equation}
     \label{eq:L}
    \mathcal{L}_\text{dist}(\hat{p}_t, \hat{p}_{s}) = -\sum_{c=1}^M \hat{p}_{t_c} \log(\hat{p}_{s_c})
\end{equation}
where
\begin{minipage}{0.46\textwidth}
  \begin{equation}
    \hat{p}_s = \softmax\left(\frac{f_t(\phi_t(x))}{\tau}\right) \label{eq:ps}
  \end{equation}
\end{minipage}
\begin{minipage}{0.46\textwidth}
  \begin{equation}
    \hat{p}_t = \softmax\left(\frac{f(\phi_{d}(x))}{\tau}\right) \label{eq:pt}
  \end{equation}
\end{minipage}

% \begin{equation*}
%     \mathcal{L}_\text{dist}(\hat{p}_t, \hat{p}_{s}) = -\sum_{c=1}^M \hat{p}_{t_c} \log(\hat{p}_{s_c})
% \end{equation*}
% %
% where,
% %
% \begin{align}
%     \hat{p}_s & = \softmax\left(\frac{f_t(\phi_t(x))}{\tau}\right) \label{eq:ps} \\
%     \hat{p}_t & = \softmax\left(\frac{f(\phi_{d}(x))}{\tau}\right) \label{eq:pt}
% \end{align}
% Here, $\tau$ is the temperature which is a hyperparameter. Consequently, the complete loss function is given by :
% %
% \begin{equation}
% \mathcal{L} = \mathcal{L}_t + \lambda \mathcal{L}_\text{dist}
% \label{eq:stage2_loss}
% \end{equation}
%
where $\lambda$ is a hyperparameter chosen from the interval $[0, 1]$.  The pseudocode for the entire training process of DeNetDM is provided in \cref{supp_sec:pseudocode}.

% % Results
\section{Experiments}
\label{sec:results}
In this section, we discuss the experimental results and analysis to demonstrate the effectiveness of DeNetDM training in debiasing. We evaluate the performance of the proposed approach by comparing it with the previous methods in debiasing, utilizing well-known datasets with diverse bias ratios, consistent with the prior works in debiasing. Additionally, we conduct an empirical study to analyze the training dynamics of DeNetDM. We also perform ablation studies to assess the effectiveness of individual components within the proposed approach.

\subsection{Experimental Setup}
\label{sec:experimental_details}
\myparagraph{Datasets:}  We evaluate the performance of DeNetDM across diverse domains using two synthetic datasets (Colored MNIST \cite{ahuja2020invariant}, Corrupted CIFAR10 \cite{hendrycks2019robustness}) and three real-world datasets (Biased FFHQ \cite{Kim_2021_ICCV}, BAR \cite{NEURIPS2020_eddc3427}) and CelebA \cite{7410782}.
In Colored MNIST (CMNIST), the digit identity is spuriously correlated with color, while in Corrupted CIFAR10 (C-CIFAR10), the texture noise corrupts the target attribute. Biased FFHQ (BFFHQ) comprises human face images from the FFHQ dataset \cite{DBLP:conf/cvpr/KarrasLA19} such that the age attribute is spuriously correlated with gender. BAR consists of human action images where six human action classes are correlated with six place attributes. We conduct experiments by varying the ratio of bias-conflicting points in the training set to demonstrate the efficacy of our approach across diverse scenarios. Following the experimental settings used by the previous works \cite{liu2023avoiding,NEURIPS2021_disentangled, 10.1007/978-3-031-19806-9_6}, we vary the ratio of bias-conflicting samples, specifically setting it at \{0.5\%, 1\%, 2\%, 5\%\} for CMNIST and C-CIFAR10, \{0.5\%\} in BFFHQ and \{1\%, 5\%\} in BAR datasets. We employ a subsampled version of CelebA as described in \cite{NEURIPS2021_de8aa43e}, maintaining the same data splits for consistency. 
 
\myparagraph{Baselines:} We compare the performance of our proposed approach to the following bias mitigation techniques; ERM \cite{DBLP:journals/tnn/Vapnik99}, GDRO \cite{sagawa2019distributionally}, LfF \cite{NEURIPS2020_eddc3427}, JTT \cite{liu2021just} , DFA \cite{NEURIPS2021_disentangled} and LC \cite{liu2023avoiding}. Among these, GDRO utilizes supervision on bias whereas LfF and JTT assumes no prior knowledge on the bais labels. DFA and LC utilizes augmentation techniques to increase diversity of minority groups. More details on the baselines are provided in \cref{supp_baselines} of the Appendix.

\begin{table*}[ht]
\caption{Testing accuracy on CMNIST and C-CIFAR10, considering diverse percentages of bias-conflicting samples. Baseline results for C-CIFAR10 are taken from \cite{liu2023avoiding}, as we employ the same experimental settings. For CMNIST, we utilize the official repositories to obtain the models. Model requirements for spurious attribute annotations (type) are indicated by \xmark~(not required) and \cmark~(required).}
\label{tab:base_results_1}
\centering
\resizebox{\textwidth}{!}{
\begin{tabular}{lcccccccccccc}
\toprule
\textbf{Methods} & \textbf{Group} && \multicolumn{4}{c}{\textbf{CMNIST}} && \multicolumn{4}{c}{\textbf{C-CIFAR10}} \\
\cmidrule(lr){4-7} \cmidrule(lr){9-12}
& \textbf{Info} && 0.5 & 1.0 & 2.0 & 5.0 && 0.5 & 1.0 & 2.0 & 5.0 \\
\midrule
Group DRO & \cmark && 59.67 & 71.33  & 76.30  & 84.40  && 33.44 & 38.30 & 45.81 & 57.32 \\
\midrule
ERM & \xmark && 35.34 (0.13) & 50.34 (0.16) & 62.29 (1.47) & 77.63 (0.13) && 23.08 (1.25) & 25.82 (0.33) & 30.06 (0.71) & 39.42 (0.64) \\
JTT & \xmark && 53.03 (3.89) & 61.68 (2.02) & 74.23 (3.21) & 85.03 (1.10) && 24.73 (0.60) & 26.90 (0.31) & 33.40 (1.06) & 42.20 (0.31) \\
LfF & \xmark && 63.39 (1.97) & 74.01 (2.21) & 80.48 (0.45) & 85.39 (0.94) && 28.57 (1.30) & 33.07 (0.77) & 39.91 (0.30) & 50.27 (1.56) \\
DFA & \xmark && 59.12 (3.15) & 71.04 (1.02) & 82.86 (2.27) & 88.29 (1.50) && 29.95 (0.71) & 36.49 (1.79) & 41.78 (2.29) & 51.13 (1.28) \\
LC & \xmark && {63.48 (5.22)} & {78.41 (1.95)} & {83.63 (1.43)} & {88.18 (1.59)} && {34.56 (0.69)} & {37.34 (1.26)} & { \bfseries 47.81 (2.00)} & {54.55 (1.26)} \\
\midrule
DeNetDM & \xmark && \bfseries 74.72 (0.99) & \bfseries 85.22 (0.76) & \bfseries 89.29 (0.51) & \bfseries 93.54 (0.22) && \bfseries 38.93 (1.16) & \bfseries 44.20 (0.77) & 47.35 (0.70) & \bfseries 56.30 (0.42) \\
\bottomrule
\end{tabular}
}
\end{table*}

\begin{table*}[ht]
\caption{Testing accuracy on BAR, BFFHQ, and CelebA. The test set for BAR and BFFHQ contains only bias-conflicting samples. Baseline method results are derived from \cite{Lim_2023_CVPR} for BAR,\cite{liu2023avoiding} for BFFHQ, and \cite{DBLP:conf/cvpr/ParkLLY23} for CelebA on the same dataset split since we utilize identical experimental settings.}
	\label{tab:base_results_2}
    \centering
    \resizebox{0.65\textwidth}{!}{
    \begin{tabular}{lcccccc}
    \toprule
    \textbf{Methods} & \textbf{Group} & \multicolumn{2}{c}{\textbf{BAR}} & \textbf{BFFHQ} & \textbf{CelebA} \\
    \cmidrule(lr){3-6}
    & \textbf{Info} & 1.0 & 5.0 & 1.0 & - \\
    \midrule
    ERM & \xmark & 57.65 (2.36) & 68.60 (2.25) & 56.7 (2.7) & 47.02 \\
    JTT & \xmark & 58.17 (3.30) & 68.53 (3.29) & 65.3 (2.5) & 76.80 \\
    LfF & \xmark & 57.71 (3.12) & 67.48 (0.46) & 62.2 (1.6) & - \\
    DFA & \xmark & 52.31 (1.00) & 63.50 (1.47) & 63.9 (0.3) & 65.26 \\
    LC & \xmark & 70.94 (1.46) & 74.32 (2.42) & 70.0 (1.4) & - \\
    \midrule
    DeNetDM (ours) & \xmark & \bfseries 73.84 (2.56) & \bfseries 79.61 (3.18) & \bfseries 75.7 (2.8) & \bfseries 81.04 \\
    \bottomrule
    \end{tabular}
    }
\end{table*}

% \begin{table*}[ht]
% \caption{Testing accuracy on BAR, BFFHQ and CelebA. The test set for BAR and BFFHQ contains only bias-conflicting samples. Baseline method results are derived from \cite{Lim_2023_CVPR} for BAR, \cite{liu2023avoiding} for BFFHQ and \cite{DBLP:conf/cvpr/ParkLLY23} for CelebA on the same dataset split since we utilize identical experimental settings.}
% 	\label{tab:base_results_2}
%     \centering
%     \resizebox{0.5\textwidth}{!}{
%     \begin{tabular}{lccccc}
%     \toprule
%     \textbf{Methods} & \textbf{Group} & \multicolumn{2}{c}{\textbf{BAR}} & \textbf{BFFHQ} \\
%     \cmidrule(lr){3-5}
%     & \textbf{Info} & 1.0 & 5.0 & 1.0 \\
%     \midrule
%     ERM & \xmark & 57.65 (2.36) & 68.60 (2.25) & 56.7 (2.7) \\
%     JTT & \xmark & 58.17 (3.30) & 68.53 (3.29) & 65.3 (2.5) \\
%     LfF & \xmark & 57.71 (3.12) & 67.48 (0.46) & 62.2 (1.6) \\
%     DFA & \xmark & 52.31 (1.00) & 63.50 (1.47) & 63.9 (0.3) \\
%     LC & \xmark & 70.94 (1.46) & 74.32(2.42) & 70.0 (1.4) \\
%     \midrule
%     DeNetDM (ours) & \xmark & \bfseries 73.84 (2.56) & \bfseries 79.61 (3.18) & \bfseries 75.7 (2.8) \\
%     \bottomrule
%     \end{tabular}
%     }
% \end{table*}

\myparagraph{Evaluation protocol:} We evaluate CMNIST and C-CIFAR10 on unbiased test sets, with target features randomly correlated to spurious features, following the evaluation protocol commonly used in prior debiasing works \cite{NEURIPS2020_eddc3427, liu2021just, NEURIPS2021_disentangled}.
Nevertheless, for BFFHQ, we do not use the unbiased test set since half of them are bias-aligned points. To ensure fair evaluation on debiasing, we adhere to previous methods \cite{liu2023avoiding, NEURIPS2021_disentangled} by exclusively utilizing a test set comprising bias-conflicting points from the unbiased test set. Notably, the BAR test set consists solely of bias-conflicting samples, posing a significant evaluation challenge. Our primary metric is accuracy, with aligned accuracy and conflicting accuracy calculated separately for some ablations on CMNIST and C-CIFAR10 (see \cref{sec:ablation studies}). Aligned accuracy is computed solely on bias-aligned data points while conflicting accuracy is determined exclusively based on the bias-conflicting points. For CelebA, we report worst-group accuracy specifically focusing on the bias-conflicting group (Blonde Hair = 0, Male = 0), which contains a substantial number of samples. We conduct five independent trials with different random seeds and report both the mean and standard deviation to ensure statistical robustness. 

\myparagraph{Implementation details:}  We perform extensive hyperparameter tuning using a small unbiased validation set with bias annotations to obtain the deep and shallow branches for all the datasets. We consistently utilize the same debiasing model architectures used by the previous methods for our target branch to ensure a fair comparison. Additionally, a linear layer is employed for the classifier for all the datasets. The additional architecture details for different datasets are as follows: \textbf{(1) CMNIST:} we use an MLP with three hidden layers for the deep branch and an MLP with a single hidden layer corresponding to the shallow branch. During the second phase of DeNetDM, we use an MLP with three hidden layers for the target branch. \textbf{(2) C-CIFAR10, BAR:} we use the ResNet-20 architecture for the deep branch and a 3-layered CNN model for the shallow branch. The target branch used in the second stage of
DeNetDM is ResNet-18. \textbf{(3) BFFHQ, CelebA:} we use the ResNet-18 architecture as the biased branch and a 4-layered CNN as the shallow branch. We also use the ResNet-18 architecture for the target branch, following the approaches of \cite{liu2023avoiding,NEURIPS2021_disentangled}.
% \end{itemize}
%
Further details on the datasets and implementation are presented in \cref{supp_sec:additional_exp_details}.  

\subsection{Evaluation Results}
We present a comprehensive comparison of DeNetDM with all the baselines described in \cref{sec:experimental_details} across varying bias conflicting ratios on CMNIST, C-CIFAR10, BFFHQ, BAR and CelebA in \cref{tab:base_results_1} and \cref{tab:base_results_2} respectively.  As evident from \cref{tab:base_results_1} and \cref{tab:base_results_2}, DeNetDM consistently outperforms all baselines across different bias ratios for CMNIST, BFFHQ, BAR and CelebA datasets. Notably, on the C-CIFAR10 dataset, DeNetDM exhibits superior performance when bias ratios are at 0.5\%, 1\%, and 5\%, and closely aligns with LC \cite{liu2023avoiding} in the case of 2\%. These findings provide evidence for the practical applicability of DeNetDM.  It is worth mentioning that the proposed approach demonstrates a significant performance enhancement across all datasets compared to Group DRO, which relies on predefined knowledge of bias. DeNetDM achieves this improvement without any form of supervision on the bias, highlighting the effectiveness of depth modulation in the debiasing.

An intriguing observation from \cref{tab:base_results_1} is that DeNetDM demonstrates better performance compared to the baselines when the bias-conflicting ratio is lower, particularly evident in the C-CIFAR10 dataset. We believe that the effectiveness of inductive bias enforced by DeNetDM in distinguishing between core and bias attributes is superior to that of LC, thereby allowing it to adeptly capture core attributes even when dealing with data points that exhibit fewer bias conflicting points. This emphasizes the applicability of DeNetDM in scenarios where the training data exhibits a significant amount of spurious correlations. Another noteworthy observation in \cref{tab:base_results_2} is that DeNetDM outperforms LC and DFA by a considerable margin across all datasets, particularly on the complex real-world datasets, BAR and BFFHQ. Both LC and DFA rely on augmentations to enhance the diversity of bias-conflicting points, whereas our approach utilizes depth modulation to efficiently capture the core attribute characteristics in the existing training data. Despite this, DeNetDM still achieves superior performance compared to LC and DFA without relying on augmentations.

\subsection{Analysis of Training Dynamics}
\begin{wrapfigure}[14]{r}{0.5\textwidth}
   \centering
   \includegraphics[width=0.5\textwidth]{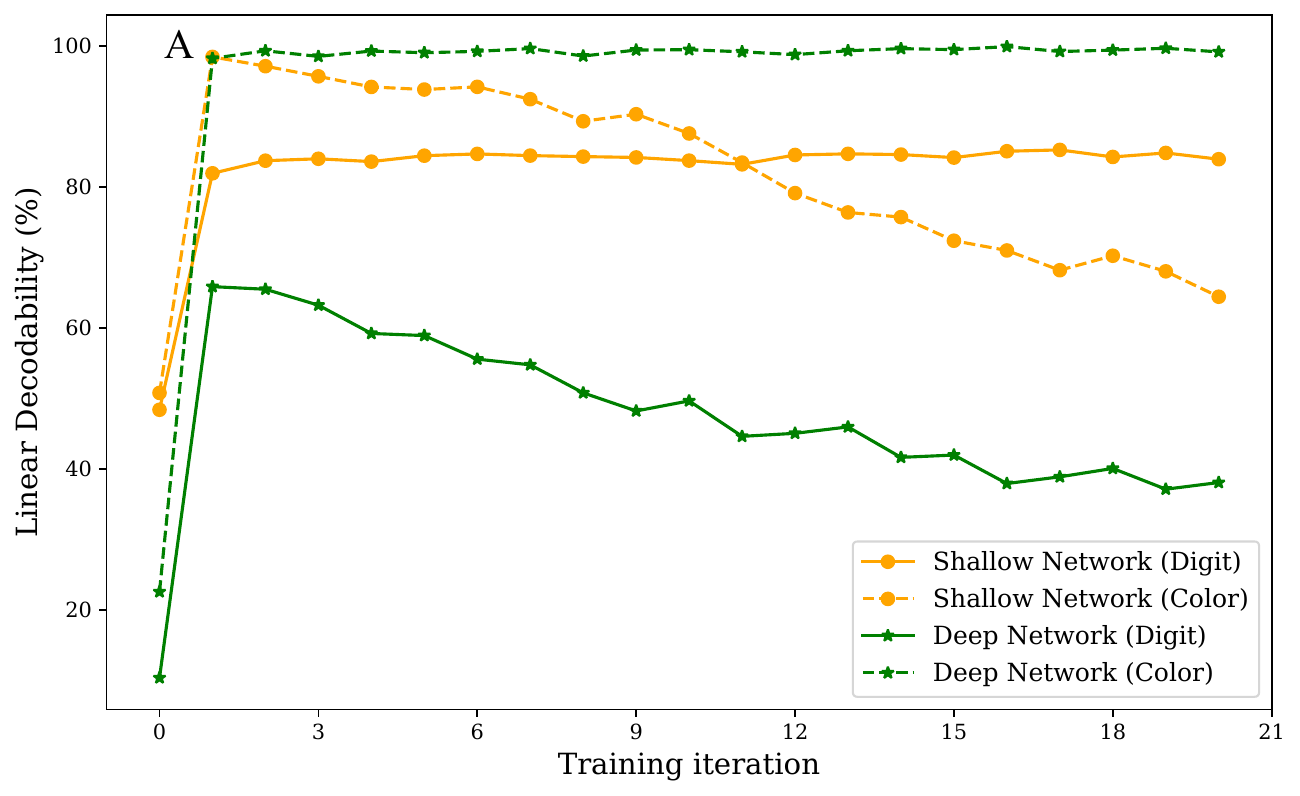}
   \caption{Early training dynamics of DeNetDM.}
   \label{fig:training_dynamics}
\end{wrapfigure}

In \cref{sec:linear_decodability_analysis}, we discussed the variability in linear decodability at various depths and its significance as a motivation for debiasing. To further validate this intuition and identify the elements contributing to its effectiveness, we delve into the training dynamics of DeNetDM during initial stages. We consider the training of Colored MNIST with 1\% skewness due to its simplicity. \Cref{fig:training_dynamics} shows how linear decodability of attributes varies across different branches of DeNetDM during training. As depicted in \cref{fig:training_dynamics}, prior to training, the deep branch demonstrates lower linear decodability for both the digit identity (core attribute) and color (bias attribute) compared to the shallow branch. As training progresses, the bias attribute, easier to learn, rapidly increases in linear decodability in both branches, labeled `A' in \cref{fig:training_dynamics}. 
\label{subsec: training dynamics analysis}
Here, the disparity in linear decodability between digit identity and color attributes becomes more pronounced in the deep branch than in the shallow one.  This distinction serves as a prior, influencing the deep branch to effectively capture the bias.  Since we employ Product of Experts technique, the deep branch becomes proficient in classification using the spurious attribute, thereby compelling the shallow branch to rely on other attributes such as digit for the classification. It is worth noting that the linear decodability of core attributes is more pronounced in the shallow branch, allowing them to capture the core attributes. Thus, the training paradigm of DeNetDM leads to a shallow branch that is robust to spurious correlations, and a deep branch that majorly relies on the biased attribute. This analysis confirms our intuition and provides empirical evidence of effective debiasing.

\subsection{Ablation Studies}
\label{sec:ablation studies}

We perform several ablation studies to evaluate different facets of DeNetDM. We scrutinize the effect of various loss components on the performance of DeNetDM.  Additionally, we explore the influence of network depth, a fundamental element of DeNetDM, and the sensitivity of DeNetDM to number of parameters which are discussed in \cref{supp_sec:additional_exp}. All the experiments are conducted on CMNIST and C-CIFAR10 datasets where the ratio of conflicting points is set to 1\%. Additional experiments and ablations are also provided in \cref{supp_sec:additional_exp}.

 \begin{table*}[h]
\caption{Ablation study of different losses used in DeNetDM on C-CIFAR10.}
\label{table:ablation-loss-components}
\centering
\resizebox{0.61\textwidth}{!}{% <------ Don't forget this %
\begin{tabular}{c c c c c c}
\toprule
{$\mathcal{L}_\text{CE}$} & {$\mathcal{L}_\text{dist}$} & { $\mathcal{L}_t$ }& {Accuracy (\%)} & {Conflicting} & {Aligned} \\
 (Stage-1) & (Stage-2) & (Stage-2) &  & Accuracy (\%) & Accuracy (\%) \\
\midrule
{\cmark} & {-} & {-} & {37.47} & {37.42} & {72.40} \\
{\cmark} & {-} & {\cmark} & {42.89} & {35.74} & {81.60} \\
{\cmark} & {\cmark}  & {-} & {42.25} & {38.34} & {68.52} \\
{\cmark} & {\cmark} & {\cmark} & {43.12} & {39.46} & {69.53} \\
% \midrule
% {1} & {CMNIST} & {\cmark} & {-} & {-} & {81.61} & {83.28} & {89.66} \\
% {2} & {CMNIST} & {\cmark} & {-} & {\cmark} & {82.96} & {81.53} & {95.85} \\
% {3} & {CMNIST} & {\cmark} & {\cmark} & {-} & {84.05} & {83.41} & {89.86} \\
% {4} & {CMNIST} & {\cmark} & {\cmark} & {\cmark} & {84.97} & {84.44} & {89.17} \\
\bottomrule
\end{tabular}
}
\end{table*}

\myparagraph{Effect of loss components:} We conduct ablation studies on C-CIFAR10 by selectively removing components to analyze their impact on the testing set accuracy as well as accuracy on bias-aligned and bias-conflicting points. The results are summarized in \cref{table:ablation-loss-components}. When considering $\mathcal{L}_\text{CE}$ alone, corresponding to the first stage of DeNetDM involving depth modulation, the model achieves 37.42\% accuracy, showing a strong ability to learn target attributes. However, introducing the second stage of DeNetDM training with $\mathcal{L}_t$ alone leads to capturing significant bias information alongside core attributes, evidenced by high accuracy on aligned points (81.60\%). When introducing $\mathcal{L}_\text{dist}$ alone, the model distills knowledge from the shallow branch obtained in the first stage, resulting in performance similar to stage 1 training. However, performing the second stage of DeNetDM training using both  $\mathcal{L}_t$ and $\mathcal{L}_\text{dist}$ prevents capturing bias, focusing more on learning core features and resulting in improved conflicting and overall accuracy. A similar trend can be observed for CMNIST dataset and the results are summarized in \cref{subsec:cmnist_loss_effect}.

\section{Conclusion}
\label{sec:conclusion}

We introduce DeNetDM, a novel debiasing method leveraging variations in linear decodability across network depths. Through extensive theoretical and experimental analysis, we uncover insights into the interplay between network architecture, attribute decodability, and training methodologies. DeNetDM employs paired deep and shallow branches inspired by the Product of Experts methodology, transferring debiasing capabilities to the desired architecture. By modulating network depths, it captures core attributes without explicit reweighting or data augmentation. Extensive experiments across various datasets, including synthetic ones like Colored MNIST and Corrupted CIFAR-10, as well as real-world datasets like Biased FFHQ and BAR, validate its robustness and superiority. Importantly, DeNetDM achieves performance comparable to supervised approaches, even without bias annotations.

% % Acknowledgements
\section{Acknowledgments}
\label{sec:ack}

Silpa Vadakkeveetil Sreelatha is partly supported by the Pioneer Centre for AI, DNRF grant number P1.

{
\small
\bibliographystyle{ieeenat_fullname}
\bibliography{main}
}

\clearpage

%%%%%%%%%%%%%%%%%%%%%%%%%%%%%%%%%%%%%%%%%%%%%%%%%%%%%%%%%%%%
% \appendix

% \appendix
% \label{sec:intro_suppliment}

\section{Appendix}
\label{sec:intro_suppliment}

In the primary text of our submission, we introduce DeNetDM, a novel debiasing framework that leverages the variation of linear decodability across network depths to effectively disentangle bias from core attributes. This technique sets a new benchmark for bias mitigation, achieving unparalleled performance without reliance on data augmentations. To ensure our manuscript's integrity, we provide an extensive appendix designed to complement the main text. This includes a series of additional experiments, expanded ablation studies, comprehensive implementation protocols, and deeper analyses of our findings. The Appendix is presented to bridge the content gap necessitated by the page constraints of the main manuscript, providing a detailed exposition of our methodology and its broader impact on the domain.
\subsection{Notations}
\label{sec:notations}
\begin{itemize}[leftmargin=*]
    \item $B$: Bias attributes
    \item $C$: Core attributes
    \item $X, Y$: Sample set (X: Inputs, Y: Labels)
    \item $X_a$: Bias-aligned points
    \item $X_b$: Bias-conflicting points
    \item $\phi$: Encoder
    \item $\phi[d]$: Encoder at depth $d$
    \item $\erank{\phi[d]}$: Effective rank of an encoder at depth $d$
    \item $\pi_i(d)$: Propagation probability of an attribute (indexed $i$) at depth $d$
    \item $\Omega^d$: Propagation probability distribution of an attribute set at layer $d$ of a neural network
    \item $\operatorname{rank}(\cdot)$: Rank of matrix
    \item $\operatorname{dim}(\cdot)$: Dimensionality of a tensor / space
    \item $\varepsilon$: Knock-off probability when transitioning from depth $d$ to $d + 1$
\end{itemize}

\subsection{Proofs}
\label{sec:proofs}

\textbf{\cref{thm:pr}} (Partition Rank):
    % Let $C$ be the random variable corresponding to the core attributes in $X$.
    % When the partitioning $X = X_a \cup X_c$ is stable \textit{wrt.} $C$, the effective rank of the bias-aligned partition is strictly lower than the effective rank of the bias-conflicting partition, \ie,
    % \begin{equation*}
    %     \erank{X_a} < \erank{X_c}
    % \end{equation*}
    % Let $C$ be the random variable corresponding to the core attributes in $X$.
    When the partitioning $X = X_a \cup X_c$ is stable \textit{wrt.} $C$, the rank of the bias-aligned partition is upper-bounded by the rank of the bias-conflicting partition, \ie,
    \begin{equation*}
        \operatorname{rank}(X_a) \leq \operatorname{rank}(X_c)
    \end{equation*}

\begin{proof}
    The theorem assumes a stable partitioning of the sample set X, \ie, in both the bias-aligned and conflicting subsets, the distribution of the core attributes are equal to that of the original sample set, \ie,
    \begin{equation*}
        P(X_a^C) = P(X_c^C) = P(X^C)
    \end{equation*}
    Under this condition, the only component in either of the subsets that determines the subset's rank should be the bias attributes, under the simplifying assumption (without loss of generality) that the attribute space is made up of only the core and the bias attributes.

    % For the bias aligned partition $X_a$, all the data points within a class have the same value for the bias attribute $B$, since it is spuriously correlated with the class label. 
    % So $B$ within a class collapses to a single scalar. Extending this across classes, $B$ over the set of all classes would map to a single vector of dimension $K$, where $K$ is the number of classes. 
    % % Assuming $\operatorname{rank}(C) > K$, the rank of $X_a$ is given by:
    % % Since $\operatorname{rank}(C) \geq 1$, we have:
    % Therefore, since the whole of $B$ in $X_a$ can be represented by a single vector orthogonal to the basis of $C$, the rank of $X_a$ is given by:
    % Assuming $\operatorname{rank}(C) > K$, the rank of $X_a$ is given by:
    % Since $\operatorname{rank}(C) \geq 1$, we have:
    % Therefore, since the whole of $B$ in $X_a$ can be represented by a single vector orthogonal to the basis of $C$, the rank of $X_a$ is given by:

    For the bias aligned partition $X_a$, all the data points within a class have very low variance within the set of values for the bias attribute $B$, since it is spuriously correlated with the class label.
    So, $B$ within a class collapses to a much lower dimensional manifold $b \subseteq B$, such that $\operatorname{rank}(b) \leq \operatorname{dim}(B)$. Extending this across classes, without loss of generality, assuming that the number of classes is higher than the variance in $B$ among the bias aligned samples, \ie, $\operatorname{rank}(b)$, $B$ over the set of all classes in $X_a$ would map to a manifold of dimensionality $\operatorname{rank}(b)$.
    Therefore, since the whole of $B$ in $X_a$ can be represented by a manifold of dimensionality $\operatorname{rank}(b)$ orthogonal to the basis of $C$, the rank of $X_a$ is given by:
    % \begin{equation*}
    %     \operatorname{rank}(X_a) = \operatorname{rank}(C) + 1
    % \end{equation*}
    \begin{equation*}
        \operatorname{rank}(X_a) = \operatorname{rank}(C) + \operatorname{rank}(b)
    \end{equation*}

    For the bias conflicting partition, since there is no correlation between the class labels and $B$, within each class, the bias attributes would require a $\operatorname{dim}(B)$ dimensional subspace independent of $C$, to be represented, since $B \perp C$. This implies that the rank of the bias conflicting points would be:
    \begin{equation*}
        \operatorname{rank}(X_c) = \operatorname{rank}(C) + \operatorname{dim}(B),
    \end{equation*}

    % As the global rank of $B$ in $X$ is strictly positive, $\operatorname{dim}(B) \geq 1$, which ultimately implies that $\operatorname{rank}(X_a) \leq \operatorname{rank}(X_c)$.
    Since we know that $b \subseteq B$, which leads to $\operatorname{rank}(b) \leq \operatorname{dim}(B)$, it is ultimately implied that $\operatorname{rank}(X_a) \leq \operatorname{rank}(X_c)$.
    
    This completes the proof of the theorem.
\end{proof}

\begin{lemma}
    \label{lemma:prop_prob_depth}
    Let $\mathcal{A} = [ A_0, A_1, ..., A_n ]$ be the attribute subspace of $X$ with increasing ranks, \ie, $\operatorname{rank}(A_0) < \operatorname{rank}(A_1) < ... < \operatorname{rank}(A_n)$,
    such that every $A \in \mathcal{A}$ is maximally and equally informative of the label $Y$, \ie, $I(A_0, Y) = I(A_1, Y) = ... = I(A_n, Y)$.
    % Then, at any given depth $d$ of a neural network, the probability of propagation $\pi_i(d)$ of an attribute $A_i$ is inversely proportional to the effective rank $\erank{\phi[d]}$ of the network at that depth, \ie,
    % \begin{equation*}
    %    \pi_i(d) = \mathcal{O} \left( \rho(\phi[d])^{-1} \right)
    % \end{equation*}
    Then, at any given depth $d$ of a neural network, the probability of propagation $\pi_i(d)$ of an attribute $A_i$ is directly proportional to the effective rank $\erank{\phi[d]}$ of the network at that depth, \ie,
    \begin{equation*}
       \pi_i(d) = \mathcal{O} \left( \rho(\phi[d]) \right)
    \end{equation*}
\end{lemma}

\begin{proof}
    Let the total rank of $\mathcal{A}$ be $R$. Consider some reference attribute $A \in \mathcal{A}$ with rank $r$. According to the results on the low rank simplicity bias \citep{huh2023simplicitybias, wang2024implicit} and deep information propagation \citep{schoenholz2017deepinfoprop}, after propagation through each layer, $\varepsilon R$ of the bases would be knocked off, resulting in a pruned version of $\mathcal{A}$. The total number of ways in which $\mathcal{A}$ can be pruned is given by
    $\begin{pmatrix} R \\ \varepsilon R \end{pmatrix}$. Also, the number of ways that $A$ features in that pruning is given by $\begin{pmatrix}
        \varepsilon R \\
        r
    \end{pmatrix}$.
    Thus, the probability of $A$ being knocked-off in layer-1 of $\phi$ is given by:
    \begin{equation*}
    %     \begin{pmatrix} R \\ \varepsilon R \end{pmatrix} /
    %     \begin{pmatrix}
    %     \varepsilon R \\
    %     r
    % \end{pmatrix}
        \begin{pmatrix} \varepsilon R \\ r \end{pmatrix} /
        \begin{pmatrix} R \\ \varepsilon R \end{pmatrix}
    = \frac{r!}{(\varepsilon R - r)! R! (1 - \varepsilon) R!}
    \end{equation*}
    Therefore, probability of survival at layer $d$:
    \begin{equation}
        \pi_i(d) = \left( 1 - \frac{r!}{\underbrace{(\varepsilon R - r)}_{a}! R! \underbrace{(1 - \varepsilon) R}_{b}!} \right)^d
        \label{eqn:prop_prob}
    \end{equation}
    Therefore, the probability of survival $\pi_i(d)$ of any attribute $A_i$ at depth $d$ increases exponentially with increasing rank $r$ of $A_i$, and decreases exponentially with the knock-off rate $\varepsilon$.
    $a$ is the part of the knocked-off basis not in $A_i$. $b$ is the part of the complete basis of $\mathcal{A}$ not affected by the first knock-off at layer 1.
    Thus, at depth $d$, $b^{d}$
    indicates the size of the subspace of $\mathcal{A}$ that survives at depth $d$, therefore being proportional to the effective rank of $\phi[d]$. Based on this, the effective rank at depth $d$ can be written as:
    % \begin{equation*}
    %     \rho(\phi[d]) = \mathcal{O}((1 - \varepsilon)^d R),
    % \end{equation*}
    % \begin{equation*}
    %     \rho(\phi[d]) = \mathcal{O}((1 - \varepsilon)^{-d} R^{-d}) = \pi_i(d),
    % \end{equation*}
    \begin{equation*}
        \pi_i(d) \propto (1 - \varepsilon)^{d} R^{d} = \mathcal{O}((1 - \varepsilon)^{d} R^{d}) = \mathcal{O}\left( \rho(\phi[d]) \right),
    \end{equation*}
    % Finally, we can write the probability of survival of $A_i$ at layer $d$ in terms of the effective rank as:
    % % \begin{align*}
    % %     \pi_i(d) &= \left( 1 - \frac{r!}{a! R! \erank{\phi[1]}!} \right)^d \\
    % %     &\implies \pi_i(d) \propto \left( \frac{1}{\rho(\phi[1])} \right)^d \\
    % %     &\implies \pi_i(d) \propto \frac{1}{\rho(\phi[d])}
    % % \end{align*}
    % % \begin{align*}
    % %     \pi_i(d) = \mathcal{O} \left( \frac{1}{(1 - \varepsilon)^d R} \right) = \mathcal{O} \left( \rho(\phi[d])^{-1} \right)
    % % \end{align*}
    % % \begin{align*}
    % %     \pi_i(d) = \mathcal{O} \left( \frac{1}{(1 - \varepsilon)^d R} \right) = \mathcal{O} \left( \rho(\phi[d]) \right)
    % % \end{align*}
    % \begin{align*}
    %     \pi_i(d) = \mathcal{O} \left( \frac{1}{(1 - \varepsilon)^d R^d} \right) = \mathcal{O} \left( \rho(\phi[d]) \right)
    % \end{align*}
    This completes the proof of the lemma.
\end{proof}

\begin{lemma}
    \label{lemma:prop_prob_rank}
    Let $\mathcal{A} = [ A_0, A_1, ..., A_n ]$ be the attribute subspace of $X$ with increasing ranks, \ie, $\operatorname{rank}(A_0) < \operatorname{rank}(A_1) < ... < \operatorname{rank}(A_n)$,
    such that every $A \in \mathcal{A}$ is maximally and equally informative of the label $Y$, \ie, $I(A_0, Y) = I(A_1, Y) = ... = I(A_n, Y)$.
    Then, at any given layer $d$ of a neural network, the propagation probability of an attribute decreases with rank, \ie,
    \begin{equation*}
        \pi_{1}(d) \geq \pi_2(d) \geq ... \geq \pi_n(d),
    \end{equation*}
    at a rate that is polynomial in the attribute rank, with degree equal to the depth, \ie,
    \begin{equation*}
        \frac{\pi_{i + 1}(d)}{\pi_i(d)} = \mathcal{O}(r^{-d})
    \end{equation*}
\end{lemma}

\begin{proof}
    Continuing from \cref{eqn:prop_prob}, we have the propagation probability of $A_i$ at depth $d$ as:
    \begin{equation*}
        \pi_i(d) = \left( 1 - \frac{r!}{\underbrace{(\varepsilon R - r)}_{a}! R! \underbrace{(1 - \varepsilon) R}_{b}!} \right)^d
    \end{equation*}
    Note that when $r$ increases, \ie, for a higher rank attribute, it leads to a drop in $a$, and in a subsequent exponential decrease in $\pi_i(d)$ as follows:
    \begin{align*}
        \pi_{i+k}(d) &= \left( 1 - \frac{(r + k)!}{(\varepsilon R - (r + k))! R! (1 - \varepsilon) R!} \right)^d \\
        & \implies \frac{\pi_{i+k}(d)}{\pi_i(d)} \leq 1 \\
        &\implies \pi_{i+k}(d) \leq \pi_i(d) \\
        &\implies \pi_{1}(d) \geq \pi_2(d) \geq ... \geq \pi_n(d),
    \end{align*}
    which proves the first part of the lemma.
    
    Now, taking the ratio of the propagation probabilities of attributes with rank $(i + k)$ and $i$ at depth $d$, we get:
    \begin{equation*}
        \frac{\pi_{i + k}(d)}{\pi_i(d)} = \mathcal{O} \left( \frac{(r+k)^d}{r^d} \right) = \mathcal{O} \left( \left( \frac{r+k}{r} \right)^d \right) = \mathcal{O}\left( \left( 1 + \frac{k}{r} \right)^d \right)
        = \mathcal{O}(k^d r^{-d})
    \end{equation*}
    For propagation on to the next layer, $k = 1$. We thus have:
    \begin{equation*}
        \frac{\pi_{i + 1}(d)}{\pi_i(d)} = \mathcal{O}(r^{-d})
    \end{equation*}
    This completes the proof of the lemma.
\end{proof}

\textbf{\cref{thm:drd}} (Depth-Rank Duality):
    Let $\mathcal{A} = [ A_0, A_1, ..., A_n ]$ be the attribute subspace of $X$ with increasing ranks, \ie, $\operatorname{rank}(A_0) < \operatorname{rank}(A_1) < ... < \operatorname{rank}(A_n)$,
    such that every $A \in \mathcal{A}$ is maximally and equally informative of the label $Y$, \ie, $I(A_0, Y) = I(A_1, Y) = ... = I(A_n, Y)$.
    Then, across the depth of the encoder $\phi$, SGD yields a parameterization that optimizes the following objective:
    \begin{equation*}
        \underbrace{\min_{\phi, f} \operatorname{\mathcal{L}}(f(\phi(X)), Y)}_\text{ERM} + \min_{\phi} \sum_d \norm{ \phi[d](\Tilde{X}) - \Omega^d \odot \mathcal{A}}_2,
        % \label{eqn:drd_opt}
    \end{equation*}
    where 
    $\mathcal{L}(\cdot, \cdot)$ is the empirical risk, $f(\cdot)$ is a classifier head, $\phi[i](\cdot)$ is the output of the encoder $\phi$ (optimized end-to-end) at depth $d$, $\norm{\cdot}_2$ is the $l^2$-norm, $\odot$ is the element-wise product,
    $\Tilde{X}$ is the $l_2$-normalized version of $X$,
    $\Omega^d = [\mathbbm{1}_{\pi_1(d)}; \mathbbm{1}_{\pi_2(d)}; ...; \mathbbm{1}_{\pi_n(d)}]$,
    $\mathbbm{1}_\pi$ is a random binary function that outputs $1$ with a probability $\pi$, and $\pi_i(d)$ is the propagation probability of $A_i$ at depth $d$ bounded as:
    % \begin{equation}
    %     \pi_i(d) = \mathcal{O}(\erank{\phi[d]}^{-1} r_i^{-d}),
    %     % \label{eqn:pos}
    % \end{equation}
    \begin{equation*}
        \pi_i(d) = \mathcal{O}\left( \erank{\phi[d]} r_i^{-d} \right),
        % \label{eqn:pos}
    \end{equation*}
    where $\erank{\phi[d]}$ is the effective rank of the $\phi[d]$ representation space, and $r_i = \operatorname{rank}(A_i)$.

\begin{proof}
    Since all $A \in \mathcal{A}$ are equally informative about the label Y, they all equally minimize $\mathcal{L}(\cdot, \cdot)$. Thus, the representations learned by $\phi$ are solely determined by the second term in the summation of \cref{eqn:drd_opt}. This means that the SGD must employ a selection mechanism to choose from the $\mathcal{A}$ that optimally utilizes the available parameters in $\mathcal{A}$.
    
    If $\phi[d]$ has sufficiently many parameters to accommodate all of $\mathcal{A}$, SGD should have no reason to discard any of them. However, a number of works that analyze the representational properties of DNNs have found that as we go deeper into a network, the effective number of dimensions available for encoding information, formally known as the effective rank and denoted as $\erank{\phi[d]}$ (effective rank of $\phi$ at depth $d$), decreases \citep{huh2023simplicitybias, wang2024implicit}. This characteristic is also known as the simplicity bias of DNNs. Given the simplicity bias, SGD must learn a parameterization for $\phi$ that optimally selects from $\mathcal{A}$ when the effective rank at a particular layer is lower than $\operatorname{rank}(\mathcal{A})$. In order to stay at the minimum of $\mathcal{L}(\cdot, \cdot)$, $\phi$ must rely on the complete basis of at least one attribute, as only partially learning an attribute would cause deviation from the minimum. So every attribute that is retained for prediction, has to be retained fully. Given this condition, the optimum choice for SGD under constrained effective ranks is thus, to choose $A \in \mathcal{A}$ in increasing order of effective ranks. In other words, the $A_0$ has the highest likelihood of getting chosen, followed by $A_1$, then $A_2$, and so on.
    
    \cref{lemma:prop_prob_depth} and \cref{lemma:prop_prob_rank} provide bounds for the quantification of the associated probabilities at a given depth, for an attribute of a given rank. Combining them, we get the propagation probability of $A_i$ at depth $d$ as:
    % \begin{equation*}
    %     \pi_i(d) = \mathcal{O}(\erank{\phi[d]}^{-1} r_i^{-d}),
    % \end{equation*}
    \begin{equation*}
        \pi_i(d) = \mathcal{O}(\erank{\phi[d]} r_i^{-d}),
    \end{equation*}
    where $r_i$ is the rank of $A_i$. We denote the distribution of $\pi$ for an attribute across a network as $\Omega_i$. Without loss of generality, assuming the retention of all attributes at depth $d-1$, we get the forward pass output at depth $d$ as:
    \begin{equation*}
        \phi[d](X) = \gamma ( W_d \cdot \phi[d-1](X) ),
    \end{equation*}
    where $W_d$ is the weight matrix at layer $d$ and gamma is a non-linearity. Under the most general setting where the elimination of attributes comes only with a decrease in the effective rank and not in the reduction in the dimensionality of the weight matrix, applying \cref{lemma:prop_prob_depth,lemma:prop_prob_rank} we obtain the survival probability of the basis corresponding to all $A \in \mathcal{A}$ in $W$ as:
    \begin{align*}
        W^d &= [\mathbbm{1}_{\pi_0(d)} W^d_0; \mathbbm{1}_{\pi_1(d)} W^d_1; ...; \mathbbm{1}_{\pi_n(d)} W^d_n] \\
        \implies W^d \cdot X' &= [\mathbbm{1}_{\pi_0(d)} W^d_0 A_0; \mathbbm{1}_{\pi_1(d)} W^d_1 A_1; ...; \mathbbm{1}_{\pi_n(d)} W^d_n A_n] \\
        &= \Omega^d \odot \mathcal{A} \cdot W
    \end{align*}
    where $\mathbbm{1}_\pi$ is a random binary function that outputs $1$ with a probability $\pi$, $\Omega^d = [\mathbbm{1}_{\pi_1(d)}; \mathbbm{1}_{\pi_2(d)}; ...; \mathbbm{1}_{\pi_n(d)}]$ , and $X' = \phi[d-1](X)$. To keep $\mathcal{L}$ at a minimum, $W^d$ must correctly activate for the informative features in $x'$, for which it must maximize $\Omega^d \odot \mathcal{A} \cdot W^d$. Now, $\Omega^d \odot \mathcal{A} \cdot W^d$ is maximized when $W^d = \Omega^d \odot \mathcal{A}$. Thus, the optimal strategy for SGD is to parameterize $W^d$ such that it captures the attributes in $\mathcal{A}$ according to the distribution $\Omega^d$. Over the full depth, the optimization objective would then be:
    \begin{equation*}
        \max_{\phi} \sum_d \Omega^d \odot \mathcal{A} \cdot \phi[d](X) \equiv \min_{\phi} \sum_d \norm{ \phi[d](\Tilde{X}) - \Omega^d \odot \mathcal{A}}_2 
    \end{equation*}
    where $\Tilde{X}$ is the $l_2$-normalized version of $X$, and the equivalence comes from the equivalence of maximizing the dot product and minimizing the $l_2$-distance of the normalized samples \citep{wiki2024metrics}.
    
    This completes the proof of the theorem.
\end{proof}

\begin{corollary}
    \label{cor:random_drd}
    Let $\mathcal{A} = [ A_0, A_1, ..., A_n ]$ be the attribute subspace of $X$ with increasing ranks, \ie, $\operatorname{rank}(A_0) < \operatorname{rank}(A_1) < ... < \operatorname{rank}(A_n)$,
    such that every $A \in \mathcal{A}$ is maximally and equally informative of the label $Y$, \ie, $I(A_0, Y) = I(A_1, Y) = ... = I(A_n, Y)$.
    Then, across the depth of a randomly initialized encoder $\phi$, the output of $\phi$ at depth $d$ follows the propagation distribution $\Omega^d$ of the attribute space $\mathcal{A}$ as:
    \begin{equation}
    \phi[d](\Tilde{X}) \propto \Omega^d \odot \mathcal{A},
    \end{equation}
    where 
    $\phi[i](\cdot)$ is the output of the encoder $\phi$ at depth $d$, $\odot$ is the element-wise product,
    $\Tilde{X}$ is the $l_2$-normalized version of $X$,
    $\Omega^d = [\mathbbm{1}_{\pi_1(d)}; \mathbbm{1}_{\pi_2(d)}; ...; \mathbbm{1}_{\pi_n(d)}]$,
    $\mathbbm{1}_\pi$ is a random binary function that outputs $1$ with a probability $\pi$, and $\pi_i(d)$ is the probability of propagation of $A_i$ of rank $r_i$ at depth $d$ bounded as:
    % \begin{equation}
    %     \pi_i(d) = \mathcal{O}(\erank{\phi[d]}^{-1} r_i^{-d}),
    % \end{equation}
    \begin{equation}
        \pi_i(d) = \mathcal{O}\left( \erank{\phi[d]} r_i^{-d} \right),
    \end{equation}
\end{corollary}

\textit{Discussion}:
    Let $\mathbb{L}$ be the space of all empirical risks $\{ \mathcal{L}_1, \mathcal{L}_2, ...\}$ over $X$. According to the No Free Lunch theorem \citep{wolpert98NoFreeLunch}, if an attribute minimizes some $\mathcal{L}_i \in \mathbb{L}$, there exists another $\mathcal{L}_j \in \mathbb{L}$ which it maximizes. So, if we consider the probability of survival of attributes in a randomly initialized network, we need to marginalize the ERM part of \cref{eqn:drd_opt}
    across the entirety of $\mathbb{L}$. Assuming an unbiased random initialization scheme, the distribution associated with $\mathbb{L}$ would be uniform (because no concrete form of empirical risk is defined, we can consider all functions $\mathcal{L} \in \mathbb{L}$ to be equally likely, under the unbiased initialization assumption) as follows:
    \begin{equation*}
        \int\limits_{\mathcal{L} \in \mathbb{L}} \operatorname{\mathcal{L}}(f(\phi(X))) \Pr(\mathcal{L}) \; d\mathcal{L},
    \end{equation*}
    where $\Pr(\mathcal{L})$ is the probability associated with the function $\mathcal{L} \in \mathbb{L}$, which can be assumed to be uniform, as argued before. Then, due to the No Free Lunch Theorem \citep{wolpert98NoFreeLunch}, the expected informativeness of all attributes in $X$ is the same, satisfying the $I(A_0, Y) = I(A_1, Y) = ... = I(A_n, Y)$ criterion in the theorem, where the nature of $Y$ is determined by the specific choice of $\mathcal{L}$.
    The remainder of the reasoning for $\phi[d](\Tilde{X}) \propto \Omega^d \odot \mathcal{A}$ is the same as the proof for $\min_{\phi} \sum_d \norm{ \phi[d](\Tilde{X}) - \Omega^d \odot \mathcal{A}}_2$ in \cref{thm:drd}.

\subsection{Equivalence with Product of Experts Framework}
\label{sec:poe_proof}

In \cref{sec:linear_decodability_analysis} of the main text, we asserted that our training methodology is derived from the Product of Experts. In this section, we elucidate this mathematically:

% Define the softmax function mapping
\begin{align*}
f &: \mathbb{R}^F \xrightarrow{\text{linear}}
 \mathbb{R}^c,  \quad  \tilde{f}(x) = \operatorname{softmax}(f(x))
\end{align*}

% Define the mapping \phi_b with a description of its domain and codomain
\begin{align*}
\phi_b &: \mathbb{R}^{C \times H \times W} \longrightarrow \mathbb{R}^F, \quad \text{where } F \text{ is the feature dimension}
\end{align*}

% Define the mapping \phi_d with a description of its domain and depth relationship to \phi_b
\begin{align*}
\phi_d &: \mathbb{R}^{C \times H \times W} \longrightarrow \mathbb{R}^F, \quad \text{such that } \operatorname{depth}(\phi_b) > \operatorname{depth}(\phi_d)
\end{align*}

% Define the loss function L
\[
\begin{aligned}
& L(x, y ; \phi_b, \phi_d)=-\sum_{c=1}^C y_c \log (\hat{p}_{\phi_b, \phi_d}^c) \quad \text{(Loss function definition)} \\
 \hat{p}_{\phi_b, \phi_d} &=\frac{\tilde{f}_c(\phi_b(x)) \cdot \tilde{f}_c(\phi_d(x))}{\sum_{c=1}^C \tilde{f}_c(\phi_b(x)) \cdot \tilde{f}_c(\phi_d(x))} \quad \text{(Product of Experts)} \\
& =\operatorname{softmax}_c(\log (\tilde{f}(\phi_b(x))) + \log (\tilde{f}(\phi_d(x)))) \quad \text{(Softmax log-sum-exp trick)} \\
& =\operatorname{softmax}_c(f(\phi_b(x)) + f(\phi_d(x))) \quad \text{(Translation invariance of softmax)} \\
& =\operatorname{softmax}_c(f(\phi_b(x) + \phi_d(x))) \quad \text{(Linearity of classifier \( f \))}
\end{aligned}
\]

We utilize $\hat{p}_{\phi_b, \phi_d}$ to compute the probabilities in DeNetDM which is the same as Equation 2 presented in the main paper.

\subsection{Pseudocode}
\label{supp_sec:pseudocode}
The pseudocode for the entire training process of DeNetDM is provided in \cref{psuedocode}.

\begin{algorithm}
    \caption{DeNetDM: Training}
    \begin{algorithmic}[1]
        \Statex \textbf{Input:} Data: $\{(x, y)_i\}_{i=1}^{N}$
        \Statex \textbf{Output:} $\phi_t, f_t$
        \Statex \textbf{Initialize:} $\phi_{t}, f_t, f ,\phi_b, \phi_d$ such that $\depth(\phi_b) > \depth(\phi_d)$
        \Repeat
            \State Fetch minibatch data $\{(x, y)_i\}_{i=1}^{K}$
            \For{$i=1$ to $K$ (in parallel) }
                \State Compute $\hat{p}$ using \eqref{eq: p_stage1} to obtain $(\hat{p}, y)_i$
            \EndFor
            % \State Update $\phi_{b}$, $\phi_d$, $f$ by SGD to minimize $\mathcal{L}(\phi_{b}, \phi_{d}, f) = \sum_{i=1}^{K} \mathcal{L}_{CE}(\hat{p}, y)_i$ where $\mathcal{L}_{CE}$ is defined by \eqref{eq:cross_entropy_stage1}
            \State Update $\phi_{b}$, $\phi_d$, $f$ by minimizing $\mathcal{L}_{CE}$ in \eqref{eq:cross_entropy_stage1} via SGD 
        \Until{Convergence} \Comment{stage1}
        \Repeat
            \State Fetch minibatch data $\{(x, y)_i\}_{i=1}^{K}$
            \For{$i=1$ to $K$ (in parallel)}
                \State Compute $\hat{p}$, $\hat{p}_s$, $\hat{p}_t$ via \eqref{eq:poe_target}, \eqref{eq:ps}, \eqref{eq:pt} respectively
            \EndFor
            \State Update $\phi_{t}$, $f_t$ by minimizing $\mathcal{L}$ in \eqref{eq:L} via SGD
        \Until{Convergence} \Comment{stage2}
    \end{algorithmic}
    \label{psuedocode}
\end{algorithm}

\subsection{Feature Decodability}
\label{sec:feature_decodability}

We utilize feature decodability to gauge the extent to which specific dataset features can be reliably decoded across models of varying depths. \cite{NEURIPS2020_71e9c662} demonstrated that the visual features can be decoded from the higher layers of untrained models. Additionally, they observed that the feature decodability from an untrained model has a significant impact in determining which features are emphasized and suppressed during the model training. Following their approach, we specifically focus on assessing the decodability of bias and core attributes from the penultimate layer of untrained models. In order to evaluate the decodability of an attribute in a dataset, we train a decoder to map the activations from the penultimate layer of a frozen, untrained model to attribute labels. The decoder comprises a single linear layer followed by a softmax activation function. The decoder is trained using an unbiased validation set associated with the dataset, where each instance is labeled according to the attribute under consideration. Subsequently, the linear decodability of the attribute, measured in accuracy, is reported on the unbiased test set. We investigate the decodability of digit and color attributes in the CMNIST dataset from MLP models with varying depths, including 3, 4, and 5 layers, and the results are depicted in \cref{fig:linear_decodability_untrained}. To investigate how feature decodability evolves during the early stages of Empirical Risk Minimization (ERM) training across networks with varying depths, we train 3-layer and 5-layer MLPs on the CMNIST dataset. Following the training, we evaluate the model's linear decodability for digit and color attributes.

\subsection{Additional Experiments}
\label{supp_sec:additional_exp}

\subsubsection{Feature Decodability on C-CIFAR10}
\label{supp_sec:decodability_cifar}

\begin{figure}
    \centering
     \includegraphics[width=0.5\textwidth]{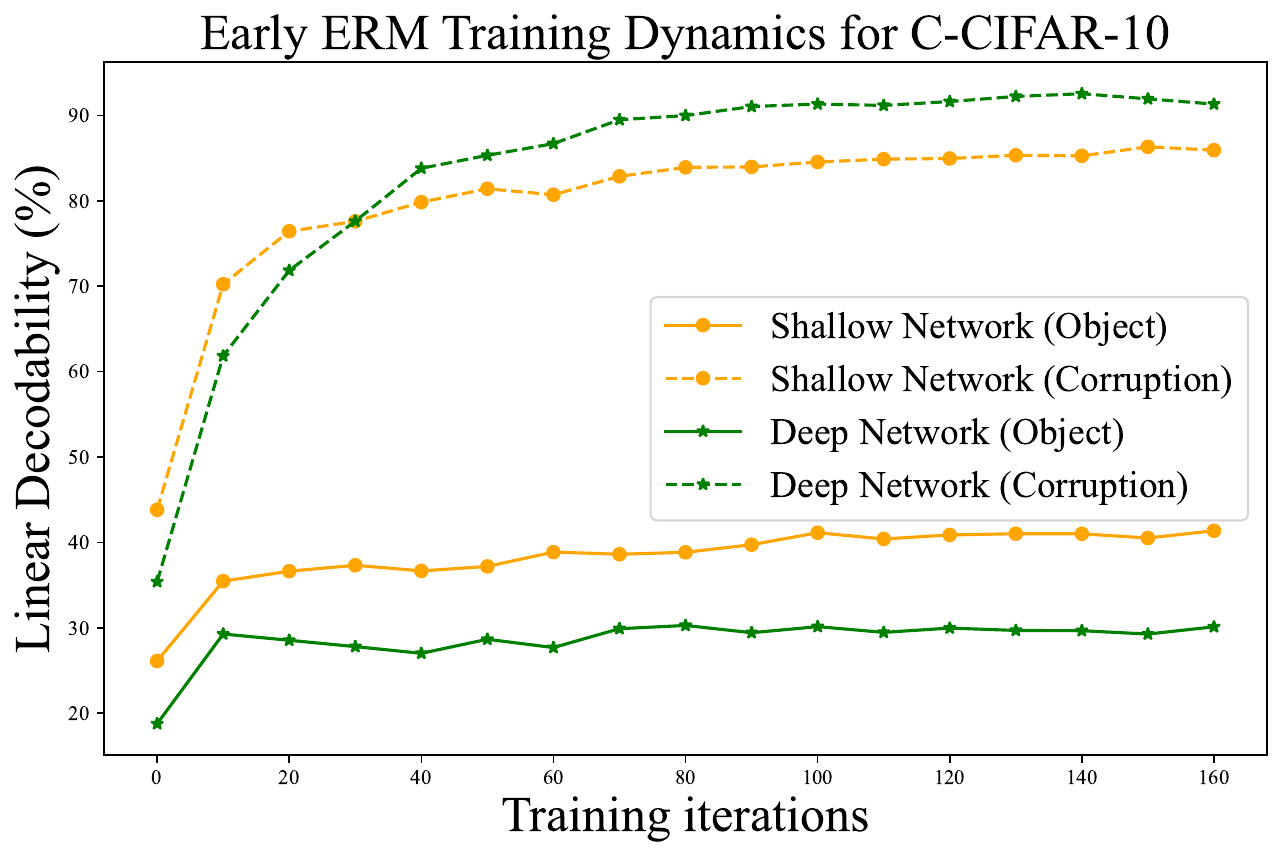}
    \caption{Early training dynamics of DeNetDM on C-CIFAR10 dataset.}
   \label{fig:training_dynamics_cifar}
\end{figure}

Analogous to \cref{fig:depth_effect} in the main paper, \cref{fig:training_dynamics_cifar} illustrates the variation in feature decodability for corruption (bias) and object (core) in the C-CIFAR10 dataset as ERM training advances. We chose ResNet 20 as the deep network and a 3-layer CNN as the shallow network since these are the architectures used for DeNetDM. The training dynamics show a similar trend to those observed in ColoredMNIST concerning bias and core attributes. As training progresses, corruption (bias) becomes highly decodable by both deep and shallow networks, with the deep branch slightly outperforming the shallow branch. However, the object attribute (core) is more decodable by the shallow network as training progresses, during the initial training dynamics. These observations align with the early training dynamics observed in CMNIST.

\subsection{Generalization to other tasks}
We evaluate the performance of DeNetDM on the CivilComments dataset \cite{pmlr-v139-koh21a}, which involves natural language debiasing. The task requires classifying online comments as toxic or non-toxic, with labels spuriously correlated with mentions of certain demographic identities. As shown in \cref{tab:worst_group_acc_civil}, our approach performs comparably to state-of-the-art methods. Due to the constrained rebuttal timeline, we just applied our model out of the box, without any reasonable hyperparameter tuning. The observations illustrate the applicability of DeNetDM to domains beyond vision.

\begin{table}[h]
\caption{Worst group accuracy (\%) comparison between different methods on CivilComments dataset.}
\label{tab:worst_group_acc_civil}
\centering
\resizebox{0.4\textwidth}{!}{
\begin{tabular}{lcc}
\toprule
\textbf{Method} & \textbf{Worst Group Acc (\%)} \\
\midrule
ERM & 58.6 (1.7) \\
JTT & 69.3 (-) \\
LfF & 58.3 (0.5) \\
LC & 70.3 (1.2) \\
DeNetDM (ours) & 68.33 (-) \\
\bottomrule
\end{tabular}
}
\end{table}

\subsubsection{Effect of depth modulation} 
To validate our hypothesis on the significance of network depth in DeNetDM, we conduct an ablation by setting the same depth for both branches and compare it with the default DeNetDM, where one branch is deeper than the other. We focus on the first stage of DeNetDM training for 5 different random seeds, reporting the averaged test accuracy on bias-aligned and bias-conflicting points for individual branches in \cref{tab:effect-depth-difference}. Branch 1 and Branch 2 in \cref{tab:effect-depth-difference} correspond to the deep and shallow branches in DeNetDM, respectively. We ignore the second stage of training since our focus was primarily on the segregation of bias and core attributes. An interesting observation is the significant standard deviation in accuracies when the branches have the same depth, observed in both datasets. This phenomenon occurs because in such a configuration, DeNetDM loses its ability to clearly distinguish between branches. This is due to the similarity in feature decodability of bias and core attributes across both branches, as discussed in \cref{sec:linear_decodability_analysis}. As a result, DeNetDM may distribute information across multiple branches or still separate core and bias attributes, but the specific branch capturing core attributes varies with different initialization.
In contrast, when depths are unequal, the deeper branch tends to focus on aligned points, disregarding conflicting ones, as seen in the test accuracies provided in \cref{tab:effect-depth-difference}. Additionally, the shallow branch emphasizes capturing core attributes, consistently enhancing conflicting accuracy. This shows the pivotal role of depth modulation in the DeNetDM framework for effectively segregating bias and core attributes.

\begin{table*}[ht]
% \centering
  \caption{Performance of DeNetDM using different network depths for the two branches of DeNetDM.}
  \label{tab:effect-depth-difference}
    \centering
    % \resizebox{0.9\textwidth}{0.3\textwidth}{
    \begin{tabular}{lccccc}
      \toprule
      Dataset & Depth & Branch & Conflicting & Aligned\\
       & (Branch 1, Branch 2) &  & Accuracy (\%) & Accuracy (\%) \\
      \midrule
      \multirow{4}{*}{CMNIST} & \multirow{2}{*}{(5, 5)} & Branch 1 & \textbf{44.94 (22.25)} & 74.85 (12.71) \\
      & & Branch 2 & 17.25 (7.89) & \textbf{88.57 (9.50)} \\
      \cmidrule{2-5}
      & \multirow{2}{*}{(5, 3)} & Branch 1 & 1.921 (0.29) & \textbf{99.92 (0.25)}\\
      & & Branch 2 & \textbf{83.17 (0.96)} & 88.25 (2.254) \\
      \midrule
       \multirow{4}{*}{C-CIFAR10} & \multirow{2}{*}{(ResNet-20, ResNet-20)} & Branch 1 & 19.54 (11.16) & 85.83 (8.19) \\
      & & Branch 2 & \textbf{24.42 (16.93)} & \textbf{86.95 (11.04)}\\
      \cmidrule{2-5}
      & \multirow{2}{*}{(ResNet-20, 3-layer CNN)} & Branch 1 & 3.0 (1.29) & \textbf{99.34 (0.47)}\\
      & & Branch 2 & \textbf{38.52 (0.99)} & 76.72 (2.19)\\
      \bottomrule
    \end{tabular}
  % }

\end{table*}

\subsubsection{Effect of loss components on CMNIST}
\label{subsec:cmnist_loss_effect}
The primary text, constrained by spatial limitations, only includes an ablation study detailing the effect of individual loss components of DeNetDM on the C-CIFAR10 dataset. However, this section extends the scope of our analysis to encompass the CMNIST dataset and the results are summarized in \cref{table:ablation-loss-components_CMNIST}. The proposed approach exhibits a similar trend as observed in the case of C-CIFAR10 (presented in \cref{sec:ablation studies}).

  \begin{table*}[ht]
\caption{Ablation study of different losses used in DeNetDM on CMNIST dataset.}
\label{table:ablation-loss-components_CMNIST}
\centering
\resizebox{0.9\textwidth}{!}{% <------ Don't forget this %
\begin{tabular}{c c c c c c}
\toprule
{$\mathcal{L}_\text{CE}$} & {$\mathcal{L}_\text{dist}$} & { $\mathcal{L}_t$ }& {Accuracy (\%)} & {Conflicting} & {Aligned} \\
 (Stage-1) & (Stage-2) & (Stage-2) &  & Accuracy (\%) & Accuracy (\%) \\
\midrule
{\cmark} & {-} & {-} & {81.61} & {83.28} & {89.66} \\
{\cmark} & {-} & {\cmark} & {82.96} & {81.53} & {95.85} \\
{\cmark} & {\cmark} & {-} & {84.05} & {83.41} & {89.86} \\
{\cmark} & {\cmark} & {\cmark} & {84.97} & {84.44} & {89.17} \\
\bottomrule
\end{tabular}
}
  \end{table*} 

\subsubsection{Depth vs. Number of parameters}
DeNetDM employs depth modulation as its principal strategy for mitigating bias. We investigate the influence of the number of parameters of both branches on DeNetDM performance. We opt for the optimal configuration of the proposed approach on C-CIFAR10 and conducted an ablation study, employing ResNet-20 ($\depth(\phi_b) = 20$) as the deep network and a 3-layer CNN ($\depth(\phi_d) = 3$) as the shallow network. We explore three scenarios where $\left | \phi_b \right | < \left | \phi_d \right |$, $\left | \phi_b \right | \approx \left | \phi_d \right |$, and $\left | \phi_b \right | > \left | \phi_d \right |$. The first stage of DeNetDM training is then performed to analyze learning in the deep and shallow models in each of the cases, and the results are presented in \cref{tab:capacity_depth}. As indicated in \cref{tab:capacity_depth}, the shallow model exhibits increased resilience to spurious correlations, while the deep model captures bias in all three cases. This suggests that DeNetDM effectively segregates bias and core attributes regardless of the number of parameters in both branches. Interestingly, a notable finding is that the shallow model exhibits better robustness against correlations when the shallow branch possesses a greater number of parameters compared to the deep model, as evident from \cref{tab:capacity_depth}.

The findings for CMNIST mirror those observed for C-CIFAR10 as presented in \cref{tab:capacity_depth_additional}: the shallow branch demonstrates robustness to spurious correlations, whereas the deep branch consistently assimilates bias irrespective of the number of parameters in both branches. These consistent patterns across datasets reinforce the efficacy of DeNetDM in distinguishing between bias and core attributes.

\begin{table*}[ht]
\caption{Ablation study on the number of parameters of deep and shallow branches in DeNetDM using C-CIFAR10 dataset.}
\label{tab:capacity_depth}
\centering
% \resizebox{0.9\columnwidth}{!}{% <------ Don't forget this %
\begin{tabular}{cccccc}
\toprule
Case & Branch &  Conflict (\%) & Align (\%)\\
\midrule
$\phi_b > \phi_d$ & $\phi_b$ & 3.08 & \textbf{96.8} \\
                 & $\phi_d$ & \textbf{29.78} & 62.61  \\
\midrule
$\phi_b \approx \phi_d$ & $\phi_b$ & 3.48 & \textbf{95.91} \\
                        & $\phi_d$ & \textbf{28.64} & 64.32  \\
\midrule
$\phi_b < \phi_d$ & $\phi_b$ & 2.04 & \textbf{99.01} \\
                 & $\phi_d$ & \textbf{39.05} & 67.68  \\
\bottomrule
\end{tabular}
% }
\end{table*}

\begin{table*}[ht]
\caption{Ablation study on the number of parameters of deep and shallow branches in DeNetDM using CMNIST dataset.}
\label{tab:capacity_depth_additional}
\centering
% \resizebox{0.5\columnwidth}{!}{% <------ Don't forget this %
\begin{tabular}{cccccc}
\toprule
Case & Branch &  Conflicting Accuracy (\%) & Aligned Accuracy (\%)\\
\midrule
                 
$\phi_b < \phi_d$ & $\phi_b$ & 11.90 & \textbf{99.93} \\
                 & $\phi_d$ & \textbf{83.89} & 88.78 &   \\
\midrule
$\phi_b \approx \phi_d$ & $\phi_b$ & 11.87 & \textbf{99.90} \\
                       & $\phi_d$ & \textbf{83.07}  & 89.09 &  \\
\midrule
$\phi_b > \phi_d$ & $\phi_b$ & 10.79 & \textbf{98.26} \\
                 & $\phi_d$ & \textbf{83.32} & 88.61 &   \\

\bottomrule
\end{tabular}
% }
\end{table*}
 
\subsubsection{Effect  of Network Depth  on DeNetDM}
In the main text, we have illustrated how the variation in network depth affects the performance of DeNetDM.We provide an in-depth analysis in this section. As observed in the first three rows of \cref{tab:performance_comparison}, as the difference in network depth of deep and shallow progressively increases, the performance of the debiased model increases monotonically. Further, when we decrease the difference in depth of shallow and deep branches (rows 3 and 4) the performance decreases to 80.42\% compared to 87.37\%. Similar performance degradation can be seen when we increase the depth of the shallow network from 4 to 6 (rows 5 and 6). Hence, DeNetDM is able to distinguish bias and core attributes better when there is a significant difference between the depths of shallow and deep branches. This aligns with the observations presented in \cref{sec:linear_decodability_analysis} of the main text (Effect of depth modulation).

\begin{table*}[ht]
\caption{Performance comparison of DeNetDM for various depths of shallow and deep branches.}
\label{tab:performance_comparison}
\centering

% Adjust the font size here if needed, for example, \small or \footnotesize
% \resizebox{0.5\columnwidth}{!}{ % Resize table to fit the column width
\begin{tabular}{cccccc}
\toprule
{Depth (Shallow, Deep)} & {Conflicting Accuracy (\%)} & {Aligned  Accuracy(\%)} \\
\midrule
(3, 4) & 72.2  & 98.33 \\
(3, 5) & 80.46 & 92.87 \\
(3, 7) & 87.37 & 93.62 \\
(6, 7) & 80.42 & 96.45 \\
(4, 8) & 91.19 & 94.62 \\
(6, 8) & 69.55 & 93.83 \\
\bottomrule
\end{tabular}
% }
\end{table*}

\subsubsection{Performance on varying bias-conflicting ratios} 
We perform experiments on the CMNIST dataset with bias-conflicting ratios of 10\% and 20\% to evaluate our method's efficacy across a broader range of ratios. The findings, presented in \cref{tab:cmnist_conflict}, show that DeNetDM performs as expected, effectively capturing core attributes in the shallow branch for varied bias ratios.

\begin{table*}
\caption{Results on CMNIST with wider bias conflicting ratios.}
\label{tab:cmnist_conflict}
\centering
% \resizebox{0.5\columnwidth}{!}{% <------ Don't forget this %
\begin{tabular}{cccccc}
\toprule
Bias ratio & Branch &  Conflicting Accuracy (\%) & Aligned Accuracy (\%)\\
\midrule 
\multirow{2}{*}{10\%} & Deep & 1.84(0.5) & \textbf{99.14(0.2)} \\
                 & Shallow & \textbf{93.12(0.8)} & 96.47(1.3) &   \\

\midrule

\multirow{2}{*}{20\%} & Deep & 3.23(2.8) & \textbf{97.93(2.1)} \\
                 & Shallow & \textbf{94.49(2.4)} & 97.51(3.4) &   \\

\bottomrule

\end{tabular}
% }
\end{table*}

\begin{table*}[ht]
  \caption{Comparison of the performance of DeNetDM using different network depths for the two branches of DeNetDM.}
  \label{tab:effect-depth}
    \centering
    \begin{tabular}{cccc}
      \toprule
      Depth (Branch 1, Branch 2) & Branch & Conflicting Accuracy (\%) & Aligned Accuracy (\%) \\
      \midrule
      \multirow{2}{*}{(ResNet-50, ResNet-32)} & Branch 1 & 3.48 (0.98) & \textbf{97.15 (2.10)} \\
                                              & Branch 2 & \textbf{30.88 (1.22)} & 81.72 (0.73) \\
      \multirow{2}{*}{(ResNet-50, ResNet-8)}  & Branch 1 & 9.38 (1.52) & \textbf{98.60 (0.86)} \\
                                              & Branch 2 & \textbf{20.32 (1.90)} & 59.94 (2.61) \\
      \bottomrule
    \end{tabular}
\end{table*}

\subsubsection{Early training dynamics in ResNet architectures}
\label{subsec:early_resnet}
We also examine the early training dynamics of ResNet-8, ResNet-32, and ResNet-50, akin to \cref{fig:training_dynamics_erm} in C-CIFAR10 dataset to assess the scalability of DeNetDM to larger ResNet models. After 200 iterations, texture (bias) decodability in all architectures neared 99\%, while core attribute decodability for ResNet-8, ResNet-32, and ResNet-50 was 18.74\%, 24.32\%, and 12.91\%, respectively. This aligns with our hypothesis that ResNet-50 would prefer texture attribute over core when paired with ResNet-8 or ResNet-32. To confirm, we tested two setups: (1) ResNet-8 and ResNet-50, and (2) ResNet-32 and ResNet-50. The results, shown in \cref{tab:effect-depth}, indicate high bias-aligned accuracy for ResNet-50 and high bias-conflicting accuracy for ResNet-8 and ResNet-32 respectively. Since ResNet-50 has lower core attribute decodability than ResNet-8 and ResNet-32, it favors bias attributes, while the shallow branches capture core attributes. This experimental results suggest DeNetDM's applicability to diverse, complex and larger models / architectures.

% \begin{table*}
%   \caption{Comparison of the performance of DeNetDM using different network depths for the two branches of DeNetDM.}
%   \label{tab:effect-depth}
%     \centering
%     \begin{tabular}{lccccc}
%       \toprule
%       Depth & Branch & Conflicting & Aligned\\
%        & (Branch 1, Branch 2) &  & Accuracy (\%) & Accuracy (\%) \\
%       \cmidrule{2-5}
%       & \multirow{2}{*}{(ResNet-50, ResNet-32)} & Branch 1 & 3.48 (0.98) & \textbf{97.15 (2.10)}\\
%       & & Branch 2 & \textbf{30.88 (1.22) } & 81.72 (0.73)\\
%       \cmidrule{2-5}
%       & \multirow{2}{*}{(ResNet-50, ResNet-8)} & Branch 1 & 9.38 (1.52) & \textbf{98.60 (0.86)}\\
%       & & Branch 2 & \textbf{20.32 (1.90)} & 59.94 (2.61)\\
%       \bottomrule
%     \end{tabular}
% \end{table*}

\subsection{Additional details}
\label{supp_sec:additional_exp_details}
In this section, we provide an in-depth discussion of various datasets used along with finer implementation details that enhance the reproducibility of our method.
\subsubsection{Datasets}
\begin{figure}[ht]
    \centering
    \includegraphics[width=0.5\columnwidth]{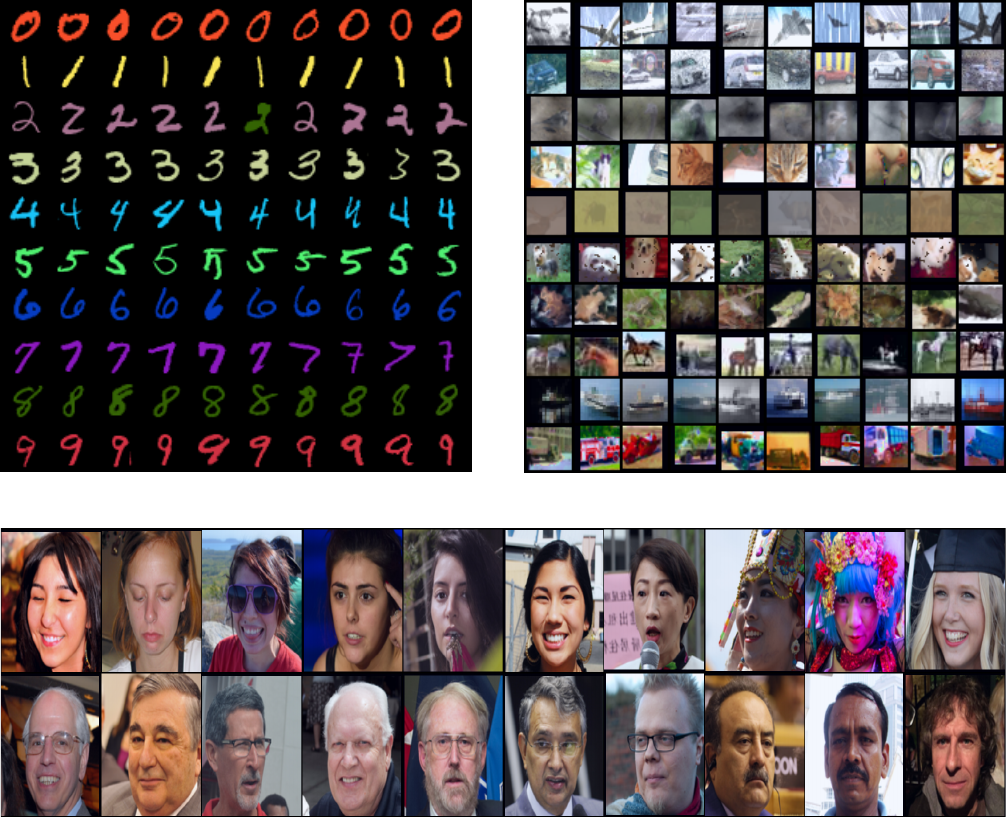}
    \caption{Samples from training data of CMNIST, Corrupted-CIFAR10 and Biased FFHQ.}
    \label{fig:training_data}
\end{figure}

We provide a detailed description of various datasets used along with a representative sample of all of them.
\begin{itemize}[leftmargin=*]
    \item \textbf{Colored MNIST(CMNIST)}: CMNIST is an adaptation of the MNIST, that introduces color variation to the images. For each digit class, the majority \( (1-\alpha) \) of the images are correlated with the corresponding color \( c_i \), with \( i \) matching the digit label \( y \). The remaining images are randomly assigned one of the other colors \( c_j \), where \( j \neq y \). The challenge of this dataset lies in identifying the digits despite the strong color bias. To incorporate color variability, a noise vector \( v \) drawn from a normal distribution is added to \( c_i \). The dataset and its characteristics are illustrated in \cref{fig:training_data}. Among multiple choices of severity, we choose the most severe corruption to simulate the worst-case scenario as done in other works.
    \item \textbf{Corrupted CIFAR10 (C-CIFAR10)}: The Corrupted CIFAR dataset represents an evolved form of the classic CIFAR set, with an emphasis on two particular features: the object depicted and the type of corruption applied. In an approach akin to that used for CMNIST, this dataset adopts an array of corruption styles, labeled from \( c_0 \), symbolizing blurring, to \( c_9 \), indicative of snow. Within each object category, a proportion \( 1-\alpha \) of the images is intentionally altered with the corruption type \( c_i \), corresponding to the object's label \( y \). The remainder of the images is processed with a randomly selected corruption type \( c_j \), chosen to ensure \( j \neq y \). In our dataset, we employ the highest degree of corruption out of the five levels outlined in the original CMNIST dataset. Illustrative samples from this dataset are demonstrated in \cref{fig:training_data}.
    \item \textbf{Biased FFHQ (BFFHQ)}\footnote{\url{https://github.com/kakaoenterprise/Learning-Debiased-Disentangled}} The BFFHQ dataset is a selectively reduced subset derived from the larger FFHQ database of facial images, with a focus on the attributes of gender and age. Gender is designated as the primary attribute of analysis, with age being the secondary attribute that could introduce bias. The gender classification is binary, encompassing male and female categories. The dataset predominantly features male images of subjects aged between 40 and 59, whereas female images are generally of subjects aged between 10 and 29. Samples that defy these age associations—such as younger male or older female subjects—are also present, countering the main age distribution.
    \item \textbf{Biased Action Recognition (BAR)} : 
The Biased Action Recognition (BAR) dataset comprises real-world images classified into six action categories, each biased towards specific locations. The chosen pairs encompass six common action-location combinations: Climbing on a Rock Wall, Diving underwater, Fishing on a Water Surface, Racing on a Paved Track, Throwing on a Playing Field, and Vaulting into the Sky. The testing set consists solely of samples that present conflicts in bias. Consequently, achieving higher accuracy results on this set indicates superior debiasing performance.
\end{itemize}

\subsubsection{Baselines}
\label{supp_baselines}
In this section, we provide a detailed overview of the baselines:
\begin{itemize}[leftmargin=*]
    \item \textbf{Empirical Risk Minimization (ERM)} \cite{DBLP:journals/tnn/Vapnik99}: Standard ERM using cross-entropy loss.
    \item \textbf{Group DRO (GDRO)} \cite{sagawa2019distributionally}: A supervised approach that utilizes group labels to identify the worst group and learn an unbiased classifier.
    \item  \textbf{Learning from Failure (LfF)} \cite{NEURIPS2020_eddc3427}: Identifies bias-conflicting points through the Generalized Cross Entropy (GCE) loss and upweighting for debiasing. 
    \item \textbf{Just Train Twice (JTT) }\cite{liu2021just}: Treats misclassified points by ERM-based classifiers as bias-conflicting and upweights them for debiasing. 
    \item  \textbf{Disentangled Feature Augmentation (DFA)} \cite{NEURIPS2021_disentangled}: Introducing feature augmentation to improve the diversity of bias-conflicting points and enhance unbiased accuracy.
    \item  \textbf{Logit Correction (LC)} \cite{liu2023avoiding}: Proposes logit correction for bias mitigation along with MixUp \cite{DBLP:conf/iclr/ZhangCDL18} inspired data augmentation for increasing diversity.
\end{itemize}

\begin{table*}[ht]
\caption{Optimal hyperparameters for the CMNIST, C-CIFAR10, BAR and BFFHQ datasets determined through extensive experimentation. The tuples represent optimal hyperparameters for Stage 1 and Stage 2, respectively.}
\label{tab:training_parameters}
\centering
\resizebox{\columnwidth}{!}{% <------ Don't forget this %
\begin{tabular}{cccc}
\toprule % from booktabs
\textbf{Parameter} & {\textbf{CMNIST}} & {\textbf{C-CIFAR10, BAR}} & {\textbf{BFFHQ}} \\
\midrule % from booktabs
Learning Rate (LR) & {$(1.0 \times 10^{-3}, 1.0 \times 10^{-3})$} & {$(1.0 \times 10^{-3}, 1.0 \times 10^{-4})$} & {$(1.0 \times 10^{-3}, 1.0 \times 10^{-4})$} \\
Batch Size         & {(64, 64)} & {(256, 256)} & {(64, 64)} \\
Momentum           & 0.9 & 0.9 & 0.9 \\
Weight Decay       & {$(1.0 \times 10^{-3}, 0)$} & {$(1.0 \times 10^{-3}, 0)$} & {(0, 0)} \\
Epochs             & {(100, 100)} & {(100, 200)} & {(10, 100)} \\
\bottomrule % from booktabs
\end{tabular}
}
\end{table*}

\subsubsection{Implementation details}
In this section, we detail the optimal hyperparameters identified for various datasets, which were instrumental in achieving the results reported in the main manuscript. The optimal hypeparameters obtained for various datasets are listed in \cref{tab:training_parameters}.Additional parameters not mentioned in \cref{tab:training_parameters} follow the default values of PyTorch.\\
\textbf{Data Augmentations:} The training phase of DeNetDM incorporated specific data augmentation techniques tailored to each dataset. For instance, the CMNIST dataset did not utilize any form of augmentation. In contrast, the C-CIFAR10 and BFFHQ datasets applied Random Horizontal Flip and random cropping, with the latter involving crops from images padded by 4 pixels. These augmentations are critical as they introduce variability into the dataset, aiding the generalization ability of the neural network.
\\
\noindent\textbf{Experimental compute:} We utilize RTX 3090 GPUs for all our experiments.\\
\noindent\textbf{Architectural Details:} Depth modulation is a critical component of our debiasing strategy. We enumerate the architecture specifics of the shallow branches tailored for each dataset below.

\textbf{CMNIST:}
\begin{lstlisting}
(shallow branch): Sequential(
  (c1): Linear(in_features=2352, out_features=100, bias=True)
  (r1): ReLU()
  (s1): MLPHiddenlayers(
    (hidden_layers): ModuleList(
      (0): Linear(in_features=100, out_features=100, bias=True)
    )
    (act): ReLU()
  )
)
\end{lstlisting}

\textbf{C-CIFAR10 and BAR:}
\begin{lstlisting}
(shallow branch): Sequential(
  (c1): Conv2d(3, 32, kernel_size=(5, 5), stride=(1, 1))
  (b1): BatchNorm2d(32, eps=1e-05, momentum=0.1, affine=True, track_running_stats=True)
  (r1): ReLU()
  (s1): MaxPool2d(kernel_size=(2, 2), stride=2, padding=0, dilation=1, ceil_mode=False)
  (c2): Conv2d(32, 64, kernel_size=(5, 5), stride=(1, 1))
  (b2): BatchNorm2d(64, eps=1e-05, momentum=0.1, affine=True, track_running_stats=True)
  (r2): ReLU()
  (s2): MaxPool2d(kernel_size=(2, 2), stride=2, padding=0, dilation=1, ceil_mode=False)
  (c3): Conv2d(64, 64, kernel_size=(5, 5), stride=(1, 1))
  (b3): BatchNorm2d(64, eps=1e-05, momentum=0.1, affine=True, track_running_stats=True)
  (r3): ReLU()
  (f1): Flatten(start_dim=1, end_dim=-1)
)
(classifier): Linear(in_features=64, out_features=10, bias=True)
(act): ReLU()
\end{lstlisting}

\textbf{BFFHQ:}
\begin{lstlisting}
(shallow branch): Sequential(
  (c1): Conv2d(3, 64, kernel_size=(7, 7), stride=(1, 1))
  (b1): BatchNorm2d(64, eps=1e-05, momentum=0.1, affine=True, track_running_stats=True)
  (r1): ReLU(inplace=True)
  (s1): MaxPool2d(kernel_size=(2, 2), stride=2, padding=0, dilation=1, ceil_mode=False)
  (c2): Conv2d(64, 128, kernel_size=(3, 3), stride=(1, 1))
  (b2): BatchNorm2d(128, eps=1e-05, momentum=0.1, affine=True, track_running_stats=True)
  (r2): ReLU(inplace=True)
  (s2): MaxPool2d(kernel_size=(2, 2), stride=2, padding=0, dilation=1, ceil_mode=False)
  (c3): Conv2d(128, 512, kernel_size=(3, 3), stride=(1, 1))
  (s3): MaxPool2d(kernel_size=(2, 2), stride=2, padding=0, dilation=1, ceil_mode=False)
  (b3): BatchNorm2d(512, eps=1e-05, momentum=0.1, affine=True, track_running_stats=True)
  (r3): ReLU(inplace=True)
  (c4): Conv2d(512, 512, kernel_size=(3, 3), stride=(1, 1))
  (b4): BatchNorm2d(512, eps=1e-05, momentum=0.1, affine=True, track_running_stats=True)
  (r4): ReLU(inplace=True)
  (a1): AdaptiveAvgPool2d(output_size=(1, 1))
  (f1): Flatten(start_dim=1, end_dim=-1)
)
\end{lstlisting}

\subsection{Limitations \& Broader Impact}
\label{supp_sec:limitations}

The primary challenge with this approach is the scalability issue when applied to a multi-bias setting. As the number of bias attributes increases, the subtle variations in linear decodability across the various branches could become so refined that accurately identifying biases may fail to achieve high fidelity. Moreover, depending on the network architecture might compel the model to depend excessively on intricate hyperparameter adjustments.

The societal impacts of identifying and mitigating biases in neural networks are extensive, resulting in fairer, more equitable, and trustworthy AI systems. Some of them are as follows :

\begin{enumerate}[leftmargin=*]
    \item Bias Mitigation in AI : contributes to more equitable AI systems by reducing the influence of spurious correlations.
    \item Societal Benefits: contributes to societal fairness by reducing biased decision-making in AI systems and potentially decreases the risk of discrimination in AI applications.
    \item Ethical AI Development: encourages transparency and accountability in AI research and deployment.
\end{enumerate}

%%%%%%%%%%%%%%%%%%%%%%%%%%%%%%%%%%%%%%%%%%%%%%%%%%%%%%%%%%%%

\newpage
\section*{NeurIPS Paper Checklist}

\begin{enumerate}

\item {\bf Claims}
    \item[] Question: Do the main claims made in the abstract and introduction accurately reflect the paper's contributions and scope?
    \item[] Answer: \answerYes{} % Replace by \answerYes{}, \answerNo{}, or \answerNA{}.
    \item[] Justification: Refer to \cref{sec:intro}.
    \item[] Guidelines:
    \begin{itemize}
        \item The answer NA means that the abstract and introduction do not include the claims made in the paper.
        \item The abstract and/or introduction should clearly state the claims made, including the contributions made in the paper and important assumptions and limitations. A No or NA answer to this question will not be perceived well by the reviewers. 
        \item The claims made should match theoretical and experimental results, and reflect how much the results can be expected to generalize to other settings. 
        \item It is fine to include aspirational goals as motivation as long as it is clear that these goals are not attained by the paper. 
    \end{itemize}

\item {\bf Limitations}
    \item[] Question: Does the paper discuss the limitations of the work performed by the authors?
    \item[] Answer: \answerYes{} % Replace by \answerYes{}, \answerNo{}, or \answerNA{}.
    \item[] Justification: Refer to \cref{supp_sec:limitations}.
    \item[] Guidelines:
    \begin{itemize}
        \item The answer NA means that the paper has no limitation while the answer No means that the paper has limitations, but those are not discussed in the paper. 
        \item The authors are encouraged to create a separate "Limitations" section in their paper.
        \item The paper should point out any strong assumptions and how robust the results are to violations of these assumptions (e.g., independence assumptions, noiseless settings, model well-specification, asymptotic approximations only holding locally). The authors should reflect on how these assumptions might be violated in practice and what the implications would be.
        \item The authors should reflect on the scope of the claims made, e.g., if the approach was only tested on a few datasets or with a few runs. In general, empirical results often depend on implicit assumptions, which should be articulated.
        \item The authors should reflect on the factors that influence the performance of the approach. For example, a facial recognition algorithm may perform poorly when image resolution is low or images are taken in low lighting. Or a speech-to-text system might not be used reliably to provide closed captions for online lectures because it fails to handle technical jargon.
        \item The authors should discuss the computational efficiency of the proposed algorithms and how they scale with dataset size.
        \item If applicable, the authors should discuss possible limitations of their approach to address problems of privacy and fairness.
        \item While the authors might fear that complete honesty about limitations might be used by reviewers as grounds for rejection, a worse outcome might be that reviewers discover limitations that aren't acknowledged in the paper. The authors should use their best judgment and recognize that individual actions in favor of transparency play an important role in developing norms that preserve the integrity of the community. Reviewers will be specifically instructed to not penalize honesty concerning limitations.
    \end{itemize}

\item {\bf Theory Assumptions and Proofs}
    \item[] Question: For each theoretical result, does the paper provide the full set of assumptions and a complete (and correct) proof?
    \item[] Answer: \answerYes{} % Replace by \answerYes{}, \answerNo{}, or \answerNA{}.
    \item[] Justification: All proofs are provided in \cref{sec:proofs}.
    \item[] Guidelines:
    \begin{itemize}
        \item The answer NA means that the paper does not include theoretical results. 
        \item All the theorems, formulas, and proofs in the paper should be numbered and cross-referenced.
        \item All assumptions should be clearly stated or referenced in the statement of any theorems.
        \item The proofs can either appear in the main paper or the supplemental material, but if they appear in the supplemental material, the authors are encouraged to provide a short proof sketch to provide intuition. 
        \item Inversely, any informal proof provided in the core of the paper should be complemented by formal proofs provided in appendix or supplemental material.
        \item Theorems and Lemmas that the proof relies upon should be properly referenced. 
    \end{itemize}

    \item {\bf Experimental Result Reproducibility}
    \item[] Question: Does the paper fully disclose all the information needed to reproduce the main experimental results of the paper to the extent that it affects the main claims and/or conclusions of the paper (regardless of whether the code and data are provided or not)?
    \item[] Answer: \answerYes{} % Replace by \answerYes{}, \answerNo{}, or \answerNA{}.
    \item[] Justification: All the experimental details required for reproducibility are covered in the \cref{sec:experimental_details,supp_sec:additional_exp_details}.
    \item[] Guidelines:
    \begin{itemize}
        \item The answer NA means that the paper does not include experiments.
        \item If the paper includes experiments, a No answer to this question will not be perceived well by the reviewers: Making the paper reproducible is important, regardless of whether the code and data are provided or not.
        \item If the contribution is a dataset and/or model, the authors should describe the steps taken to make their results reproducible or verifiable. 
        \item Depending on the contribution, reproducibility can be accomplished in various ways. For example, if the contribution is a novel architecture, describing the architecture fully might suffice, or if the contribution is a specific model and empirical evaluation, it may be necessary to either make it possible for others to replicate the model with the same dataset, or provide access to the model. In general. releasing code and data is often one good way to accomplish this, but reproducibility can also be provided via detailed instructions for how to replicate the results, access to a hosted model (e.g., in the case of a large language model), releasing of a model checkpoint, or other means that are appropriate to the research performed.
        \item While NeurIPS does not require releasing code, the conference does require all submissions to provide some reasonable avenue for reproducibility, which may depend on the nature of the contribution. For example
        \begin{enumerate}
            \item If the contribution is primarily a new algorithm, the paper should make it clear how to reproduce that algorithm.
            \item If the contribution is primarily a new model architecture, the paper should describe the architecture clearly and fully.
            \item If the contribution is a new model (e.g., a large language model), then there should either be a way to access this model for reproducing the results or a way to reproduce the model (e.g., with an open-source dataset or instructions for how to construct the dataset).
            \item We recognize that reproducibility may be tricky in some cases, in which case authors are welcome to describe the particular way they provide for reproducibility. In the case of closed-source models, it may be that access to the model is limited in some way (e.g., to registered users), but it should be possible for other researchers to have some path to reproducing or verifying the results.
        \end{enumerate}
    \end{itemize}

\item {\bf Open access to data and code}
    \item[] Question: Does the paper provide open access to the data and code, with sufficient instructions to faithfully reproduce the main experimental results, as described in supplemental material?
    \item[] Answer: \answerYes{} % Replace by \answerYes{}, \answerNo{}, or \answerNA{}.
    \item[] Justification: Source code is provided in \href{https://github.com/kadarsh22/DeNetDM}{https://github.com/kadarsh22/DeNetDM}.
    \item[] Guidelines:
    \begin{itemize}
        \item The answer NA means that paper does not include experiments requiring code.
        \item Please see the NeurIPS code and data submission guidelines (\url{https://nips.cc/public/guides/CodeSubmissionPolicy}) for more details.
        \item While we encourage the release of code and data, we understand that this might not be possible, so “No” is an acceptable answer. Papers cannot be rejected simply for not including code, unless this is central to the contribution (e.g., for a new open-source benchmark).
        \item The instructions should contain the exact command and environment needed to run to reproduce the results. See the NeurIPS code and data submission guidelines (\url{https://nips.cc/public/guides/CodeSubmissionPolicy}) for more details.
        \item The authors should provide instructions on data access and preparation, including how to access the raw data, preprocessed data, intermediate data, and generated data, etc.
        \item The authors should provide scripts to reproduce all experimental results for the new proposed method and baselines. If only a subset of experiments are reproducible, they should state which ones are omitted from the script and why.
        \item At submission time, to preserve anonymity, the authors should release anonymized versions (if applicable).
        \item Providing as much information as possible in supplemental material (appended to the paper) is recommended, but including URLs to data and code is permitted.
    \end{itemize}

\item {\bf Experimental Setting/Details}
    \item[] Question: Does the paper specify all the training and test details (e.g., data splits, hyperparameters, how they were chosen, type of optimizer, etc.) necessary to understand the results?
    \item[] Answer: \answerYes{} % Replace by \answerYes{}, \answerNo{}, or \answerNA{}.
    \item[] Justification: Refer to \cref{sec:experimental_details,supp_sec:additional_exp_details}
.    \item[] Guidelines:
    \begin{itemize}
        \item The answer NA means that the paper does not include experiments.
        \item The experimental setting should be presented in the core of the paper to a level of detail that is necessary to appreciate the results and make sense of them.
        \item The full details can be provided either with the code, in appendix, or as supplemental material.
    \end{itemize}

\item {\bf Experiment Statistical Significance}
    \item[] Question: Does the paper report error bars suitably and correctly defined or other appropriate information about the statistical significance of the experiments?
    \item[] Answer: \answerYes{} % Replace by \answerYes{}, \answerNo{}, or \answerNA{}.
    \item[] Justification: All the experiments are conducted for 5 different random seeds and we report the mean and standard deviation. 
    \item[] Guidelines:
    \begin{itemize}
        \item The answer NA means that the paper does not include experiments.
        \item The authors should answer "Yes" if the results are accompanied by error bars, confidence intervals, or statistical significance tests, at least for the experiments that support the main claims of the paper.
        \item The factors of variability that the error bars are capturing should be clearly stated (for example, train/test split, initialization, random drawing of some parameter, or overall run with given experimental conditions).
        \item The method for calculating the error bars should be explained (closed form formula, call to a library function, bootstrap, etc.)
        \item The assumptions made should be given (e.g., Normally distributed errors).
        \item It should be clear whether the error bar is the standard deviation or the standard error of the mean.
        \item It is OK to report 1-sigma error bars, but one should state it. The authors should preferably report a 2-sigma error bar than state that they have a 96\% CI, if the hypothesis of Normality of errors is not verified.
        \item For asymmetric distributions, the authors should be careful not to show in tables or figures symmetric error bars that would yield results that are out of range (e.g. negative error rates).
        \item If error bars are reported in tables or plots, The authors should explain in the text how they were calculated and reference the corresponding figures or tables in the text.
    \end{itemize}

\item {\bf Experiments Compute Resources}
    \item[] Question: For each experiment, does the paper provide sufficient information on the computer resources (type of compute workers, memory, time of execution) needed to reproduce the experiments?
    \item[] Answer: \answerYes{} % Replace by \answerYes{}, \answerNo{}, or \answerNA{}.
    \item[] Justification: Please refer to \cref{supp_sec:additional_exp_details}.
    \item[] Guidelines:
    \begin{itemize}
        \item The answer NA means that the paper does not include experiments.
        \item The paper should indicate the type of compute workers CPU or GPU, internal cluster, or cloud provider, including relevant memory and storage.
        \item The paper should provide the amount of compute required for each of the individual experimental runs as well as estimate the total compute. 
        \item The paper should disclose whether the full research project required more compute than the experiments reported in the paper (e.g., preliminary or failed experiments that didn't make it into the paper). 
    \end{itemize}
    
\item {\bf Code Of Ethics}
    \item[] Question: Does the research conducted in the paper conform, in every respect, with the NeurIPS Code of Ethics \url{https://neurips.cc/public/EthicsGuidelines}?
    \item[] Answer: \answerYes{} % Replace by \answerYes{}, \answerNo{}, or \answerNA{}.
    \item[] Justification: The proposed approach conforms with the NeurIPS Code of Ethics.
    \item[] Guidelines:
    \begin{itemize}
        \item The answer NA means that the authors have not reviewed the NeurIPS Code of Ethics.
        \item If the authors answer No, they should explain the special circumstances that require a deviation from the Code of Ethics.
        \item The authors should make sure to preserve anonymity (e.g., if there is a special consideration due to laws or regulations in their jurisdiction).
    \end{itemize}

\item {\bf Broader Impacts}
    \item[] Question: Does the paper discuss both potential positive societal impacts and negative societal impacts of the work performed?
    \item[] Answer: \answerYes{} % Replace by \answerYes{}, \answerNo{}, or \answerNA{}.
    \item[] Justification: Please refer to \cref{supp_sec:limitations}
    \item[] Guidelines:
    \begin{itemize}
        \item The answer NA means that there is no societal impact of the work performed.
        \item If the authors answer NA or No, they should explain why their work has no societal impact or why the paper does not address societal impact.
        \item Examples of negative societal impacts include potential malicious or unintended uses (e.g., disinformation, generating fake profiles, surveillance), fairness considerations (e.g., deployment of technologies that could make decisions that unfairly impact specific groups), privacy considerations, and security considerations.
        \item The conference expects that many papers will be foundational research and not tied to particular applications, let alone deployments. However, if there is a direct path to any negative applications, the authors should point it out. For example, it is legitimate to point out that an improvement in the quality of generative models could be used to generate deepfakes for disinformation. On the other hand, it is not needed to point out that a generic algorithm for optimizing neural networks could enable people to train models that generate Deepfakes faster.
        \item The authors should consider possible harms that could arise when the technology is being used as intended and functioning correctly, harms that could arise when the technology is being used as intended but gives incorrect results, and harms following from (intentional or unintentional) misuse of the technology.
        \item If there are negative societal impacts, the authors could also discuss possible mitigation strategies (e.g., gated release of models, providing defenses in addition to attacks, mechanisms for monitoring misuse, mechanisms to monitor how a system learns from feedback over time, improving the efficiency and accessibility of ML).
    \end{itemize}
    
\item {\bf Safeguards}
    \item[] Question: Does the paper describe safeguards that have been put in place for responsible release of data or models that have a high risk for misuse (e.g., pretrained language models, image generators, or scraped datasets)?
    \item[] Answer: \answerNA{} % Replace by \answerYes{}, \answerNo{}, or \answerNA{}.
    \item[] Justification: The paper poses no such risks.
    \item[] Guidelines:
    \begin{itemize}
        \item The answer NA means that the paper poses no such risks.
        \item Released models that have a high risk for misuse or dual-use should be released with necessary safeguards to allow for controlled use of the model, for example by requiring that users adhere to usage guidelines or restrictions to access the model or implementing safety filters. 
        \item Datasets that have been scraped from the Internet could pose safety risks. The authors should describe how they avoided releasing unsafe images.
        \item We recognize that providing effective safeguards is challenging, and many papers do not require this, but we encourage authors to take this into account and make a best faith effort.
    \end{itemize}

\item {\bf Licenses for existing assets}
    \item[] Question: Are the creators or original owners of assets (e.g., code, data, models), used in the paper, properly credited and are the license and terms of use explicitly mentioned and properly respected?
    \item[] Answer: \answerYes{} % Replace by \answerYes{}, \answerNo{}, or \answerNA{}.
    \item[] Justification: The paper uses existing public datasets and models and are properly credited wher ever required throughout the paper.
    \item[] Guidelines:
    \begin{itemize}
        \item The answer NA means that the paper does not use existing assets.
        \item The authors should cite the original paper that produced the code package or dataset.
        \item The authors should state which version of the asset is used and, if possible, include a URL.
        \item The name of the license (e.g., CC-BY 4.0) should be included for each asset.
        \item For scraped data from a particular source (e.g., website), the copyright and terms of service of that source should be provided.
        \item If assets are released, the license, copyright information, and terms of use in the package should be provided. For popular datasets, \url{paperswithcode.com/datasets} has curated licenses for some datasets. Their licensing guide can help determine the license of a dataset.
        \item For existing datasets that are re-packaged, both the original license and the license of the derived asset (if it has changed) should be provided.
        \item If this information is not available online, the authors are encouraged to reach out to the asset's creators.
    \end{itemize}

\item {\bf New Assets}
    \item[] Question: Are new assets introduced in the paper well documented and is the documentation provided alongside the assets?
    \item[] Answer: \answerNA{} % Replace by \answerYes{}, \answerNo{}, or \answerNA{}.
    \item[] Justification: The paper does not release new assets.
    \item[] Guidelines:
    \begin{itemize}
        \item The answer NA means that the paper does not release new assets.
        \item Researchers should communicate the details of the dataset/code/model as part of their submissions via structured templates. This includes details about training, license, limitations, etc. 
        \item The paper should discuss whether and how consent was obtained from people whose asset is used.
        \item At submission time, remember to anonymize your assets (if applicable). You can either create an anonymized URL or include an anonymized zip file.
    \end{itemize}

\item {\bf Crowdsourcing and Research with Human Subjects}
    \item[] Question: For crowdsourcing experiments and research with human subjects, does the paper include the full text of instructions given to participants and screenshots, if applicable, as well as details about compensation (if any)? 
    \item[] Answer: \answerNA{} % Replace by \answerYes{}, \answerNo{}, or \answerNA{}.
    \item[] Justification: The paper does not involve crowdsourcing nor research with human subjects.
    \item[] Guidelines:
    \begin{itemize}
        \item The answer NA means that the paper does not involve crowdsourcing nor research with human subjects.
        \item Including this information in the supplemental material is fine, but if the main contribution of the paper involves human subjects, then as much detail as possible should be included in the main paper. 
        \item According to the NeurIPS Code of Ethics, workers involved in data collection, curation, or other labor should be paid at least the minimum wage in the country of the data collector. 
    \end{itemize}

\item {\bf Institutional Review Board (IRB) Approvals or Equivalent for Research with Human Subjects}
    \item[] Question: Does the paper describe potential risks incurred by study participants, whether such risks were disclosed to the subjects, and whether Institutional Review Board (IRB) approvals (or an equivalent approval/review based on the requirements of your country or institution) were obtained?
    \item[] Answer: \answerNA{} % Replace by \answerYes{}, \answerNo{}, or \answerNA{}.
    \item[] Justification: The paper does not involve crowdsourcing nor research with human subjects.
    \item[] Guidelines:
    \begin{itemize}
        \item The answer NA means that the paper does not involve crowdsourcing nor research with human subjects.
        \item Depending on the country in which research is conducted, IRB approval (or equivalent) may be required for any human subjects research. If you obtained IRB approval, you should clearly state this in the paper. 
        \item We recognize that the procedures for this may vary significantly between institutions and locations, and we expect authors to adhere to the NeurIPS Code of Ethics and the guidelines for their institution. 
        \item For initial submissions, do not include any information that would break anonymity (if applicable), such as the institution conducting the review.
    \end{itemize}

\end{enumerate}

\end{document}